\documentclass[10pt,journal,compsoc]{IEEEtran}

\ifCLASSOPTIONcompsoc
  \usepackage[nocompress]{cite}
\else
  \usepackage{cite}
\fi
\ifCLASSINFOpdf
   \usepackage[pdftex]{graphicx}
\else
\fi
\usepackage{amsmath}
\usepackage{acronym}
\usepackage[noend]{algorithmic}
\usepackage[linesnumbered,ruled,vlined]{algorithm2e}

\usepackage{array}

\usepackage{mdwmath}
\usepackage{mdwtab}

\usepackage{eqparbox}

\ifCLASSOPTIONcompsoc
 \usepackage[caption=false,font=footnotesize,labelfont=sf,textfont=sf]{subfig}
\else
 \usepackage[caption=false,font=footnotesize]{subfig}
\fi

\usepackage{stfloats}
\ifCLASSOPTIONcaptionsoff
 \usepackage[nomarkers]{endfloat}
\let\MYoriglatexcaption\caption
\renewcommand{\caption}[2][\relax]{\MYoriglatexcaption[#2]{#2}}
\fi
\usepackage{url}
\usepackage{soul, color}
\definecolor{hhhh}{RGB}{51,107,158}
\newcommand\MYhyperrefoptions{bookmarks=true,bookmarksnumbered=true,
pdfpagemode={UseOutlines},plainpages=false,pdfpagelabels=true,
colorlinks=true,linkcolor={black},citecolor={black},urlcolor={black},allcolors={hhhh},
pdftitle={Learning Large Neighborhood Search for
Vehicle Routing in Airport Ground Handling},
pdfsubject={Airport Ground Handling},
pdfauthor={ianan Zhou, Yaoxin Wu, Zhiguang Cao, Wen Song, Jie Zhang, and Zhenghua Chen},
pdfkeywords={Data-driven Optimization, Learning to Optimize, Airport Ground Handling, Large Neighborhood Search, Graph Neural
Network, Deep Learning}}
\ifCLASSINFOpdf
\usepackage[\MYhyperrefoptions,pdftex]{hyperref}
\else
\usepackage[\MYhyperrefoptions,breaklinks=true,dvips]{hyperref}
\usepackage{breakurl}
\fi

\usepackage{cite}
\usepackage{ulem}
\usepackage{amsmath,amssymb,amsfonts}
\usepackage{booktabs}
\usepackage{multirow}

\begin{document}
%
\title{Learning Large Neighborhood Search for Vehicle Routing in Airport Ground Handling}
%
%
%
%

\author{Jianan Zhou, Yaoxin Wu, Zhiguang Cao, Wen Song, Jie Zhang, and Zhenghua Chen
\IEEEcompsocitemizethanks{\IEEEcompsocthanksitem Jianan Zhou and Jie Zhang are with the School of Computer Science and Engineering, Nanyang Technological University, Singapore. (emails: jianan004@e.ntu.edu.sg, zhangj@ntu.edu.sg) \protect
\IEEEcompsocthanksitem Yaoxin Wu is with the Department of Information Systems, Faculty of Industrial Engineering and Innovation Sciences, Eindhoven University of Technology, Netherlands. (emails: wyxacc@hotmail.com) \protect
\IEEEcompsocthanksitem Zhiguang Cao is with the School of Computing and Information Systems, Singapore Management University, Singapore. (email: zhiguangcao@outlook.com) \protect
\IEEEcompsocthanksitem Wen Song is with the Institute of Marine Science and Technology, Shandong University, China. (email: wensong@email.sdu.edu.cn)\protect
\IEEEcompsocthanksitem Zhenghua Chen is with the Institute for Infocomm Research (I2R), Agency for Science Technology and Research (A*STAR), Singapore. (email: chen0832@e.ntu.edu.sg) \protect
\IEEEcompsocthanksitem Corresponding authors: Yaoxin Wu, Wen Song.}
}

%
%

\markboth{IEEE Transactions on Knowledge and Data Engineering}%
{Shell \MakeLowercase{\textit{et al.}}: Bare Advanced Demo of IEEEtran.cls for IEEE Computer Society Journals}
%



\IEEEtitleabstractindextext{%
\begin{abstract}
Dispatching vehicle fleets to serve flights is a key task in airport ground handling (AGH). Due to the notable growth of flights, it is challenging to simultaneously schedule multiple types of operations (services) for a large number of flights, where each type of operation is performed by one specific vehicle fleet. To tackle this issue, we first represent the operation scheduling as a complex vehicle routing problem and formulate it as a mixed integer linear programming (MILP) model. Then given the graph representation of the MILP model, we propose a learning assisted large neighborhood search (LNS) method using data generated based on real scenarios, where we integrate imitation learning and graph convolutional network (GCN) to learn a destroy operator to automatically select variables, and employ an off-the-shelf solver as the repair operator to reoptimize the selected variables. Experimental results based on a real airport show that the proposed method allows for handling up to 200 flights with 10 types of operations simultaneously, and outperforms state-of-the-art methods. Moreover, the learned method performs consistently accompanying different solvers, and generalizes well on larger instances, verifying the versatility and scalability of our method.
\end{abstract}

}

\maketitle

\IEEEdisplaynontitleabstractindextext

%
\IEEEpeerreviewmaketitle

\begin{figure}[!t]
\centering
\includegraphics[scale=0.12]{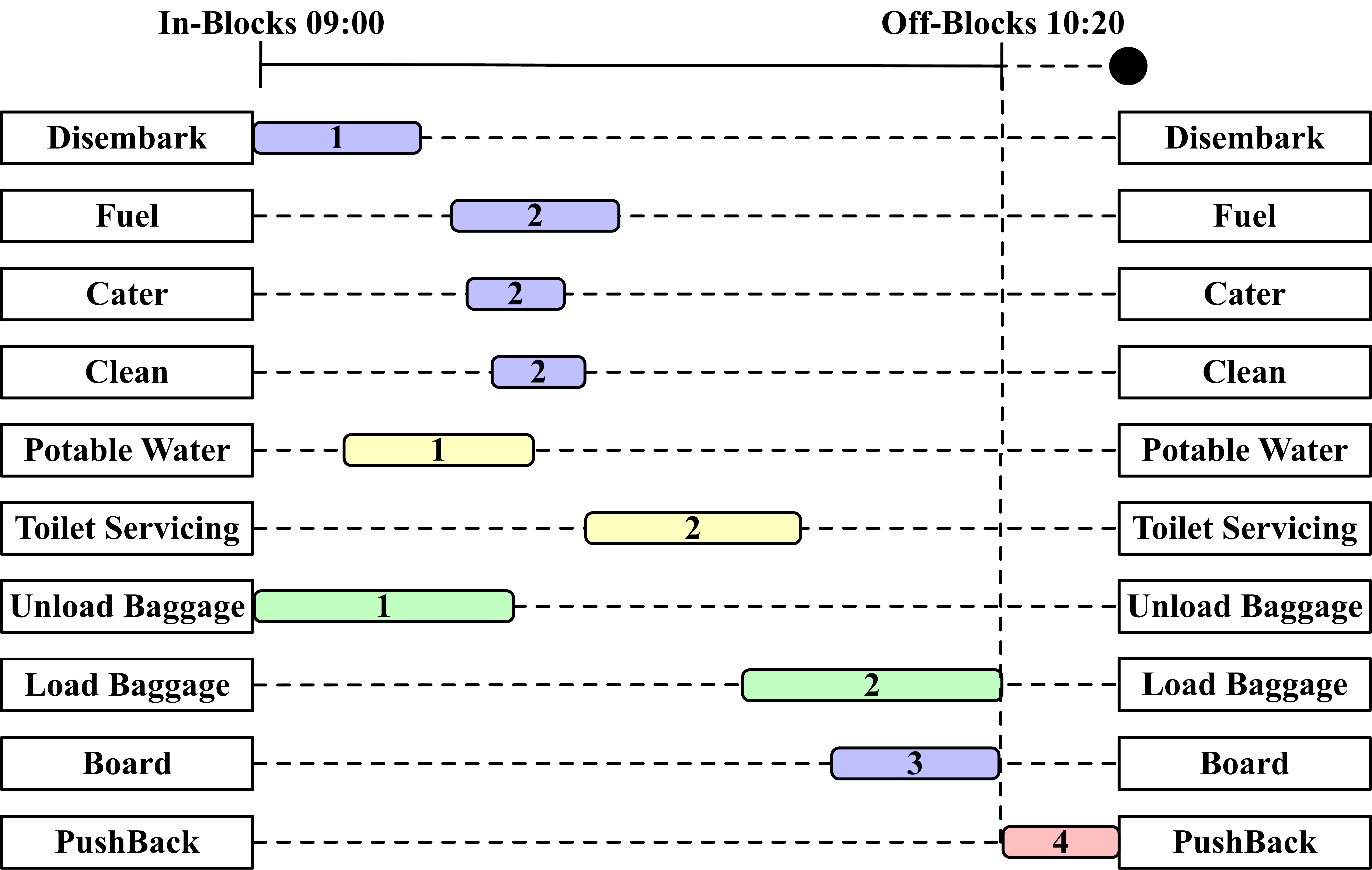}
\caption{An example of AGH operations and their precedence. Operations with the same color have precedence relation and the one with a lower value needs to perform earlier, e.g., fueling, catering and cleaning with the same priority precede boarding.}
\label{fig_0}
\vspace{-4mm}
\end{figure}

\ifCLASSOPTIONcompsoc
\IEEEraisesectionheading{\section{Introduction}\label{sec:introduction}}
\else
\section{Introduction}
\label{sec:introduction}
\fi
\IEEEPARstart{D}{ue} to the growing demands for air travelling, the airports worldwide are becoming much busier and more crowded, causing severe flight delays and substantial economic loss. To counteract the undesirable impacts, one attainable measure is ameliorating the efficiency of airport ground handling (AGH) based on the historical data, which hinges on how to efficiently dispatch the vehicle fleets for serving the flights by performing the needed operations when they land in the stand for turnarounds. Dispatching the vehicle fleets could be naturally formulated as a vehicle routing problem (VRP), however, the one in AGH is hard to be solved. On the one hand, it always needs to serve a large number of flights at a time and each needs multiple (types of) operations, such as disembarking, refueling, cleaning and so on. On the other hand, each (type of) operation may need a fleet of vehicles to perform, and complex precedence relations between different operations are always required, e.g., the one in Fig.~\ref{fig_0}. Therefore, solving the VRP in AGH is nontrivial yet challenging for the aviation industry. Although some commercialized exact solvers based on the branch and bound framework (or its variants) have been developed to optimally solve combinatorial optimization problems including VRP, e.g., CPLEX~\cite{cplex2009v12} and Gurobi~\cite{gurobi}, they may cost prohibitively long time when directly solving the routing problems in AGH given their intractability.   

A practical alternative to tackle such hard problems is the use of heuristic methods. They have the potential to deliver sub-optimal solutions in reasonable computation time. In those methods, certain hand-crafted rules are always designed to cope with the hard constraints so that the computation could be expedited when searching a solution of satisfactory quality. Following this principle, a number of heuristic methods have been attempted to handle the simple variants of VRP in AGH, which only consider a single operation, i.e., de-icing~\cite{norin2012scheduling}, towing~\cite{du2014planning}, catering~\cite{ho2010solving}, fueling~\cite{du2008aco}, trailering~\cite{zhou2018research}, and baggage loading and unloading~\cite{guo2020scheduling}, respectively. 
Although some success has been achieved, they fail to consider those operations simultaneously and their precedence relation, which may render them less effective for solving the real world routing problems in AGH. Among this line of research, a notable work considers a relatively hard variant of VRP in AGH with multiple operations~\cite{padron2016bi}. To solve this problem, it first leverages constraint programming (CP) to decompose the whole problem into VRP with time window (VRPTW) for each operation, and then exploits the large neighborhood search (LNS) to solve the respective sub-problems. Recently, another work exploits genetic algorithm (GA) to tackle the decomposed routing problems in AGH, which encodes the sequence of vehicle allocation as chromosome~\cite{liu2021scheduling}. However, these methods may need massive domain knowledge for AGH and heavy hand-engineering for the key designs, e.g., how to properly decompose the original problems; how to effectively design operators in LNS or genetic algorithms. 

On the other hand, the industrial counterpart might be keen on the off-the-shelf solvers for a fast deployment rather than designing complex heuristics for every new problem, although directly using them often suffers from poor performance on large or complex problems. Hence, it would be more desirable and practical to learn improving the performance of those solvers from their externals, e.g., the mixed integer linear programming (MILP) solvers like CPLEX~\cite{cplex2009v12} or OR-Tools~\cite{ortools}, for solving VRP in AGH. Ideally, the proposed method will, 1) learn to automatically decompose the problems to better take the advantage of the solvers, without much domain knowledge for AGH; 2) also circumvent the design of labor intensive rules for the resulting subproblems, given the direct use of solvers with default settings.

To this end, in this paper, we present a learning based LNS method to directly solve the MILP model of the VRP in AGH, which can tackle 200 flights and 10 operations (e.g. the one in Fig.~\ref{fig_0}). Specifically, we first model the problem as a multiple-fleet VRP with various constraints pertaining to capacity, time windows, precedence, etc., and formulate it as a MILP model. Then, in our LNS method, we leverage an imitation learning and a bipartite graph convolutional network (GCN) to learn the \emph{destroy} operator so that it would automatically select the decision variables in the MILP model at each step, which would be subsequently reoptimized by adopting an off-the-shelf solver, i.e., CPLEX or OR-Tools, as the \emph{repair} operator. In doing so, the original hard problem is automatically decomposed into a series of sub-MILP problems with less complexity, which is supposed to considerably save the computation. Our contributions are threefold as follows.

\begin{itemize}
  \item [1)] 
  We propose a LNS method to solve the routing problems in AGH, where we exploit destroy operator to select decision variables in the MILP model and repair operator to reoptimize them. Thus, the original problem is decomposed into sub-problems.
  \item [2)] 
  We leverage the imitation learning and bipartite GCN to learn the destroy operator given the graph representation of the MILP model, which circumvents the handed-crafted rules in the conventional heuristic methods, and also enables desirable versatility for the repair operator.
  \item [3)] 
  We conduct extensive experiments under realistic settings of the well-known CHANGI airport. Results show that our method allows for handling up to 200 flights and 10 types of operations with superior performance to the conventional ones. 
\end{itemize}

The remainder of this paper is organized as follows. Section~\ref{sec:related_work} reviews existing works related to ours. Section~\ref{sec:MILP} introduces the vehicle routing problem in AGH with multiple operations, and formulates it as a MILP model. Section \ref{sec:methodology} elaborates our learning based LNS method. Section \ref{sec:experiments} presents empirical results and analysis. The conclusion and future work are finally stated in Section \ref{sec:conclusion}.

\section{Related Works}
\label{sec:related_work}
In this section, we review the existing works pertaining to airport ground handling (AGH), and typical LNS methods for solving VRP and its variants, respectively. Recent studies that combine LNS with learning methods are also discussed. 

\subsection{Airport Ground Handling}
AGH, as one of the main airport activities, plays an important role in raising the operational efficiency of airports and the service quality of air transportation. Most of early attempts on the AGH problem focused on scheduling vehicles for performing a single (type of) operation. Norin et al. modeled the scheduling problem for the de-icing operation as a VRPTW, with the objective to reduce the delay of flights and traveling distance of de-icing vehicles~\cite{norin2009integrating,norin2012scheduling}. They proposed a greedy randomized adaptive search procedure (GRASP), which was regarded as a simplified variant of LNS, and delivered superior results in comparison with other types of greedy heuristics. Du et al. coped with a scheduled problem for the towing operation by leveraging the column generation heuristic to solve a MILP problem \cite{du2014planning}. Ho and Leung evaluated the performance of both tabu search and simulated annealing for solving the scheduling problem with respect to catering operation \cite{ho2010solving}. Du et al. modeled the scheduling problem for fueling operation as a multi-objective VRPTW and solved it by an ant colony optimization (ACO) heuristic~\cite{du2008aco}. Zhou et al. tackled the trailer scheduling problem by developing a multi-target scheduling model with rolling windows~\cite{zhou2018research}. Guo et al. performed the scheduling of baggage transport vehicles by leveraging both population diversity and fitness in a genetic algorithm~\cite{guo2020scheduling}.
Han et al. described the scheduling of ferry vehicles as a VRPTW, and designed three heuristic algorithms to solve the problem~\cite{han2022abi}.
Nevertheless, the above methods ignore the underlying interplay between multiple operations and thus may deliver inferior solutions to AGH. On the other hand, the global optimization regarding AGH with multiple operations has received relatively less attention. Among the existing works, Padr{\'o}n et al. proposed to simultaneously schedule multiple AGH operations at an airport \cite{padron2016bi}, where constrain programming (CP) was adopted to assign time windows to vehicle fleets and then the resulting VRPTW for each fleet was solved by LNS, with the objective to reduce the waiting time of flights and the duration of turnarounds. This method is further ameliorated in terms of computational efficiency in their subsequent work~\cite{padron2019improved}.  
Liu et al. considered scheduling each type of vehicle (fleet) as a separate sub-problem, which was solved by the genetic algorithm (GA)~\cite{liu2021scheduling}. Different from ours, it optimized both the number of vehicles and time cost, and attained superior performance to the multiobjective evolutionary algorithm based on decomposition (MOEAD) and the particle swarm optimization (PSO) algorithm. Recently, Zhang et al. used GA to solve the scheduling problem of supporting vehicles, which is similar to AGH but ignores the travel time of vehicles between aircraft. The method was evaluated with less than 10 aircraft, which is far from the ones in practical situations~\cite{zhang2022optimization}.

Besides the AGH, there exists literature on tackling other operational issues in airports. For example,
Yıldız et al. developed an object detection system to monitor and analyze the ground service actions from video frame sequences~\cite{yildiz2022turnaround}. Alomar and Tolujevs developed a simulation model to generate event streams and movements of the mobile objects, aiming to test control techniques in the airport~\cite{alomar2017optimization}.
Another line of research analyzes safety issues in airports, which are captured by the interplay between air and ground vehicles or the use of civil drones~\cite{shvetsov2021analysis,shvetsova2021ensuring}.
Other less related works for manpower scheduling or airside operation research can be found in~\cite{gok2020robust,gok2020simheuristic,andreatta2014efficiency,ng2018review}.

\subsection{Large Neighborhood Search}
LNS is a classic metaheuristic based on local search, which has demonstrated competitive performance for solving various VRPs~\cite{shaw1998using}, such as the pickup and delivery problem \cite{ropke2006adaptive}, the dial-a-ride problem \cite{jain2011large}, and the capacitated VRP (CVRP) \cite{ribeiro2012adaptive, sacramento2019adaptive, xiao2019evolutionary}. With destroy and repair procedures, LNS decomposes the original intractable problem into a series of subproblems, which are iteratively solved to fast improve the solution. However, in LNS, a number of rules have to be manually designed for the respective problems, e.g., heuristics in operators, the choice of operators, destroy degree, etc. Regarding the routing problems in AGH with multiple operations, it could be represented as a multiple-fleet VRP with complex constraints, which is more computationally intractable than the common VRPs as mentioned above. Designing rules for such problem usually needs substantial domain knowledge. Existing works~\cite{padron2016bi, padron2019improved} handle a similar variant of this problem by decomposing it into VRPTW for each respective fleet using constraint programming, and then sequentially solve them by LNS. In contrast, we directly leverage LNS to decompose the MILP model of the problem, by exploiting imitation learning to learn desirable behaviors of the destroy operator. We also apply existing solvers as repair operators to save manual efforts in designing meticulous rules otherwise. In doing so, our method greatly reduces the need for much domain knowledge and tuning work in comparison to classic LNS methods, which not only accelerates the algorithmic development but also promotes the eventual performance.



On the other hand, some recent attempts have been performed to learn destroy or repair operators in LNS. Hottung et al. utilized an attention model to learn repair operators for standard CVRP \cite{hottung2019neural}. Gao et al. adopted recurrent neural network (RNN) and graph attention network (GAT) to learn a reinsertion operator for CVRP with time windows \cite{gao2020learn}. However, neural networks involved in these methods are only applicable to a specified type of VRP. Instead of learning problem-dependent operators, Song et al. proposed to destroy a solution by selecting variables at each step in LNS, and reoptimize them by an existing solver \cite{song2020general}. This method is relatively general and relies on less domain knowledge. However, it fixes the number of the variables that need to be reoptimized to a constant value, which is determined by prior trials for each problem. Also, they trained fully-connected neural networks with incidence matrices of same-sized instances as features, which limits its scalability. In contrast, our method allows reoptimizing a flexible number of variables at each step, and the GCN based policy network is less sensitive to problem scale. Besides, we would like to note that, our method is also different from the automatic neighborhood design, which usually groups entities (e.g., customers, commodities) from the MILP model and then optimizes the resulting small portions \cite{ghiani2015model,adamo2020learn}. Most of this line of works pursue a good feasible solution with one-step clustering by manually designing metrics to compute distances between the entities. In contrast, we aim to improve solutions sequentially with imitation learning and enable the training of GCN to automatically select variables for optimization at each step.

\section{Problem Statement and MILP Model}
\label{sec:MILP}
In this section, we describe the vehicle routing problem in AGH for performing multiple operations and also present its MILP formulation.

Pertaining to our problem, each type of operation is conducted by a vehicle fleet. Therefore, we describe it as a multiple-fleet VRP with constraints of \emph{capacity}, \emph{time windows} and \emph{precedence}. Formally, the problem is defined on an undirected graph $G=(\Omega,E)$ with the node set $\Omega=\{0,1,\cdots,n, \dot{n}\}$ and edge set $E=\{(i,j)|i,j\in \Omega; i\neq j\}$. In the node set, $0$ and $\dot{n}$ represent the initial and final depot node (refer to the same one), and $\Omega^* =\Omega \setminus\{0, \dot{n}\}$ represent flights with demand $q_i,i\in \Omega^*$, where we set $q_0=q_{\dot{n}}=0$. Each edge $(i,j)\in E$ bears an associated cost $c_{ij}^k$ for operation $k$, where we set $c_{0 \dot{n}}^k=0, \forall k\in \mathcal{K}$. The heterogeneous vehicle fleets $\mathcal{K}=\{1,\cdots,K\}$ are responsible for different operations, and $k_1\prec k_2$ ($k_1, k_2\in\mathcal{K}$) means the operation by $k_1$ precedes that of $k_2$. Each fleet $k\in \mathcal{K}$ has $V^k$ homogeneous vehicles $\mathcal{V}^k=\{1,\cdots,V^k\}$ with the same capacity $Q^k$. Regarding a vehicle in the fleet $k$, $s_{i}^k$ denotes its service duration for flight $i$, and $s_0^k=s_{\dot{n}}^k=0$; $t_{ij}^k$ denotes the travel time from flight $i$ to $j$, and $t_{0\dot{n}}^k=0$; 
$[a_{i}^k, b_{i}^k]$ represents the time window for servicing flight $i$. Naturally, the problem entails two classes of decision variables, 1) the binary variable $x_{ijv}^k$, which is equal to $1$ if a vehicle $v$ in fleet $k$ travelled from flight $i$ to $j$, and $0$ otherwise; 2) the real variable $T_{iv}^k$, which represents the start time for servicing flight $i$ by vehicle $v$ in fleet $k$. Accordingly, we formulate the AGH problem as follows,
\begin{align}
  \label{eq1} \text{min.} \quad & \sum_{k\in \mathcal{K}}\sum_{v\in \mathcal{V}^k}\sum_{(i,j)\in E} c_{ij}^kx_{ijv}^k\\
  \label{eq2} \text{s.t.} \quad & \sum_{i\in \Omega}\sum_{v\in \mathcal{V}^k}x_{ijv}^k=1, \forall j\in \Omega^*, k\in \mathcal{K} \\
  \label{eq3} & \sum_{i\in \Omega\setminus \{\dot{n}\}} \hspace{-2mm} x_{iuv}^k= \hspace{-2mm} \sum_{j\in \Omega\setminus\{0\}} \hspace{-3mm} x_{ujv}^k, \forall u\in \Omega^*, v\in \mathcal{V}^k, k\in \mathcal{K}
\end{align}
\begin{align}
  \label{eq4} & \sum_{j\in \Omega^*}\sum_{v\in \mathcal{V}^k}x_{0jv}^k \leq V^k, \forall k\in \mathcal{K} \\
  \label{eq5} & \sum_{j\in \Omega^*}\sum_{v\in \mathcal{V}^k}x_{0jv}^k = \sum_{i\in \Omega^*}\sum_{v\in \mathcal{V}^k}x_{i\dot{n}v}^k, \forall k \in \mathcal{K} \\
  \label{conflict} & \sum_{i\in \Omega\setminus\{0\}}\hspace{-1mm} \sum_{v\in \mathcal{V}^k}\hspace{-1mm} x_{i0v}^k=\hspace{-1mm} \sum_{j\in \Omega\setminus\{\dot{n}\}} \hspace{-1mm} \sum_{v\in \mathcal{V}^k}x_{\dot{n}jv}^k=0, \forall k\in \mathcal{K} \\    
  \label{capacity} & \sum_{i\in \Omega^*}q_i \sum_{j\in \Omega} x_{ijv}^k\leq Q^k, \forall v\in \mathcal{V}^k, k \in \mathcal{K} \\
  \label{eq8} & x_{ijv}^k(T_{iv}^k +s_i^k +t_{ij}^k - T_{jv}^k) \leq 0, \forall v\in \mathcal{V}^k, k \in \mathcal{K} \\
  \label{eq9} & a_i^k \leq T_{iv}^k \leq b_i^k, \forall v\in \mathcal{V}^k, k \in \mathcal{K} \\
  \label{eq10} & T_{iv}^{k_1} + s_i^{k_1} \leq T_{iv}^{k_2}, \forall k_1, k_2\in\mathcal{K}, k_1 \prec k_2\\
  \label{eq11} & x_{ijv}^k\in\{0,1\}, \forall (i,j)\in E, v\in \mathcal{V}^k, k \in \mathcal{K} \\
  \label{eq12} & T_{iv}^k\geq 0, \forall i\in \Omega^*, v\in \mathcal{V}^k, k \in \mathcal{K}.
\end{align}
The objective function in Eq. (\ref{eq1}) aims to minimize the total cost (distance). Eq. (\ref{eq2}) ensures that each flight is only served by one vehicle in each fleet. Eq. (\ref{eq3}) ensures the continuity of the route. Eq. (\ref{eq4}) ensures that the number of used vehicles will not exceed the available vehicles in each fleet. Eq. (\ref{eq5}) ensures the equivalence between the number of vehicles leaving and returning to the depot. Eq. (\ref{conflict}) ensures that all routes are from the initial depot node to final depot node. Eq. (\ref{capacity}) defines capacity constraints in each fleet. Eq. (\ref{eq8}) and (\ref{eq9}) ensure that the start time of a service is within the respective time windows. Eq. (\ref{eq10}) imposes that the operation by a fleet must precede that of another if they have precedence relation, as illustrated in Fig.~\ref{fig_0}. Since Eq. (\ref{eq8}) is nonlinear, we transform it as follows,
\begin{equation}
\label{eq13} T_{iv}^k +s_i^k +t_{ij}^k - T_{jv}^k \leq C(1-x_{ijv}^k), \forall (i,j)\in E, k \in \mathcal{K}   
\end{equation}
where $C$ is a large constant, and we set it to $10^6$. In doing so, we model the problem as the form of a mixed integer linear programming (MILP).

We would like to mention that the presented formulation is more complicated than most of conventional VRPs such as CVRP, VRPTW, etc. \cite{feng2020explicit,duan2021robust}. We not only tackle practical constraints in AGH, i.e., time windows, capacity or precedence constraints, but also deal with heterogeneous operations for a large volume of aircraft. We also note that the problem definition is not a mere extension of conventional VRPs but well reflects the complex context in practical AGH. All constraints in our formulation have their own physical meaning. Meanwhile, the AGH has its own exclusive attributes, e.g., it often requires decisions on both routes and start times of operations, which is much more practical and challenging than conventional VRPs. Solving AGH with such a complex context is fairly challenging. 
Thus, designing hand-crafted heuristics to solve this problem might be intractable, which is usually accompanied with substantial domain knowledge and trial-and-error efforts. In contrast, we turn to automatically learn LNS policy to solve the MILP model, which is expected to save large amounts of labour and be more friendly to non-expert users.


\begin{figure*}[!t]
\centering
\includegraphics[width=7.0in]{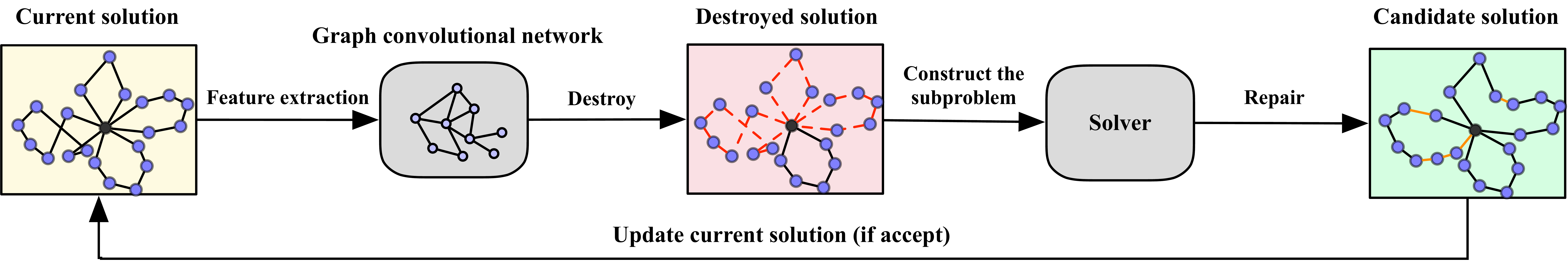}
\vspace{-1mm}
\caption{\textbf{An overview of the learning based LNS framework.} Given an instance and the current solution, the trained policy selects groups of integer decision variables associated to certain vehicles, and then the repair operator (e.g. CPLEX) solves the sub-MILP, where the selected variables are reoptimized and others are fixed to the values in the current solution. Once a new solution is computed, the current solution is updated according to an acceptance criterion. The process is repeated until a termination condition is met.}
\label{fig_1}
\vspace{-2mm}
\end{figure*}

\section{Methodology}
\label{sec:methodology}
In this section, we present the learning based LNS method to solve the routing problem in AGH. We first introduce the general LNS framework which is characterized by a destroy operator and repair operator. Then we design a simple yet efficient heuristic destroy operator which would be used to yield demos for the subsequent imitation learning. Finally, given the graph representation of our MILP model, we exploit the imitation learning and graph neural network to guide the destroy operator to automatically select variables, which will be reoptimized by the repair operator. In doing so, the original problem is decomposed into a series of sub-problems, and the heavy computation would be considerably mitigated. The proposed LNS framework with the learned destroy operator is overviewed in Fig. \ref{fig_1}.

\subsection{The LNS Framework}
As a local search based metaheuristic, LNS defines its solution neighborhood by destroy and repair operators. Formally, given an instance $p$ of a combinatorial optimization problem with a minimization objective function, the set of feasible solutions is denoted as $X(p)$. We aim to find a solution $x^* \in X(p)$ such that $c(x^*) \leq c(x)$, $\forall x \in X(p)$, where $c(\cdot)$ is the objective function. 
Generally, LNS solves the problem in an iterative manner. In each iteration, the destroy operator $\mathcal{D}$ destructs part of the current solution $\tilde{x}$, and the repair operator $\mathcal{R}$ rebuilds the destroyed solution to obtain a new solution $\tilde{x}'$. Since stochasticity exists in both destroy and repair operators, the neighborhood $N(\tilde{x}) \subseteq X(p)$ of the solution $\tilde{x}$ is defined as the set of solutions that can be derived by first applying $\mathcal{D}$ and then $\mathcal{R}$, so that $\tilde{x}'\in N(\tilde{x})$. The new solution $\tilde{x}'$ will replace the current solution according to the acceptance criterion, such as a simple scheme of accepting only improved solutions or the one in simulated annealing \cite{ropke2006adaptive,schrimpf2000record}. The above process is repeated until the termination condition is satisfied. The initial solution $\tilde{x}_0$ is obtained by either simple heuristics or exact solver with short runtime. The incumbent solution $x_b$ is updated along the solving process of LNS.



In our method, we leverage LNS to directly solve the MILP model in section~\ref{sec:MILP}. Specifically, assuming that $\mathcal{X}$ denotes the set of decision variables $x_{ijv}^k$, we employ the destroy operator in each iteration to select a subset of variables $\mathcal{X}_r$ for reoptimization, while fixing the other variables $\mathcal{X}\setminus \mathcal{X}_r$ to the values in the current solution $\tilde{x}$. A natural way for reoptimization is to use an off-the-shelf solver as the repair operator, e.g., CPLEX, especially given that the resulting problem has been scaled down. Obviously, our proposed LNS framework brings two major benefits, 1) we target on variable-level destroy operator in the context of MILP, and it eschews massive problem-specific manual rules; 2) the original problem is decomposed into a series of sub-problems of MILP, and it allows tractable computation for state-of-the-art exact solvers. Accordingly, the pseudo-code of our LNS framework is summarized in Algorithm~\ref{LNS}.


\begin{algorithm}[!t]
  \caption{Formulation based LNS Framework}
  \label{LNS}
	\begin{algorithmic}[1]
		\STATE Input: an instance of MILP problem $p$ with initial solution $\tilde{x}_0$; number of iterations $n$; destroy and repair operators $\mathcal{D}$, $\mathcal{R}$; percentage of decision variables to destroy $\mathcal{D}_{d}$.
		\STATE $x_b = \tilde{x} = \tilde{x}_0$
        \STATE $best\_obj = c(\tilde{x}_0)$
		\FOR{$t=1,\cdots,n$}
		\STATE $\mathcal{X}_r = \mathcal{D}(p, \tilde{x}, \mathcal{D}_{d})$
		\STATE $\tilde{x}'$ = $\mathcal{R}(p, \tilde{x}, \mathcal{X}_r)$
		\IF{$accept(\tilde{x}', \tilde{x})$}
		\STATE $\tilde{x} = \tilde{x}'$
		\IF{$c(\tilde{x}') < best\_obj$}
		\STATE $x_b = \tilde{x}'$
  		\STATE $best\_obj = c(\tilde{x}')$
		\ENDIF
		\ENDIF
		\ENDFOR
		\RETURN $x_b, \tilde{x}', \mathcal{X}_r$
	\end{algorithmic}
\end{algorithm}


\subsection{Heuristic Destroy Operator}
\label{subsec:destroy}
To provide demonstrations (demos) for the learning based destroy operator in section~\ref{subsec:learning}, we first design a heuristic destroy operator in consideration of \textit{destroy degree} and \textit{stochasticity}. The destroy degree is a key factor for a destroy operator, which is defined as $\mathcal{D}_d=|\mathcal{X}_r|/|\mathcal{X}|$ with $|\cdot|$ denoting the cardinality of a set. It determines the percentage of decision variables to select for reoptimization, and controls the size of the resulting sub-MILP at each step of LNS. If the subset of the selected variables is too small, the repair operator may fail to sufficiently explore the solution space. If the subset is too large, the sub-problem would be almost the same to the original problem, which may lead to prohibitively heavy computation. Besides, the stochasticity is typically required for a destroy operator to avoid the stagnation, i.e., the situation where almost the same subsets of variables are optimized in consecutive steps with the solution improved negligibly. Keeping both the destroy degree and stochasticity in mind, we propose a \emph{Vehicle-random} destroy operator to solve the MILP model in section \ref{sec:MILP}, which selects the integer decision variables related to the randomly chosen vehicle(s) given the destroy degree, i.e., $\mathcal{X}_r = \{x_{ijv}^k|k\in \mathcal{K}_s\subset \mathcal{K}, v\in\mathcal{V}_s^k \subseteq \mathcal{V}^k, \forall i,j \in \Omega\}$, where $\mathcal{K}_s$ is the indices of selected fleets and $\mathcal{V}_s^k$ is the indices of selected vehicles in the fleet $k$. 
All the continuous decision variables $T_i^k$ will be reoptimized together with the selected integer variables $\mathcal{X}_r$ by the repair operator, since continuous ones require much less computation. Although \emph{Vehicle-random} destroy operator is simple and intuitive, we could apply it multiple times to an instance and retrieve the best one, which would be adequately good to serve as demonstrations for the imitation learning.


Note that most of existing LNS methods rely on the relatedness or cost on each node (i.e., flights in our AGH problem) to design operators for specific VRPs~\cite{ropke2006adaptive}. Normally, they need substantial domain knowledge or manual efforts, and may not be directly applied to select variables from the mathematical programming formulation. In contrast, both our heuristic and learning based destroy operators have favorable potential to alleviate those issues.



\begin{figure*}[!t]
\centering
\includegraphics[width=7.0in]{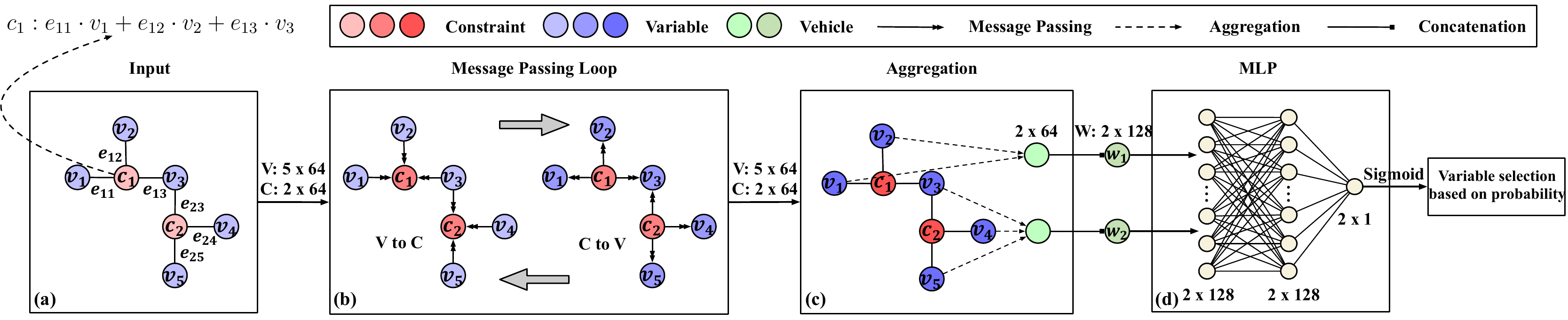}
\vspace{-1mm}
\caption{\textbf{An illustration of policy network $\pi_\theta$ with 5 variables and 2 constraints.} (a) The state of constraints and variables is represented as a bipartite graph; (b) The message passing from variables $\mathcal{V}$ ($\mathcal{C}$) to constraints $\mathcal{C}$ ($\mathcal{V}$) updates the constraint (variable) embeddings; (c) The variable embeddings are aggregated into vehicle nodes $\mathcal{W}$ by mean pooling over variables related to the same vehicle; (d) The forward passing of the aggregated embeddings is further processed through the MLP. The embeddings for each kind of nodes are being updated as the color deepens.}
\label{fig_2}
\vspace{-2mm}
\end{figure*}

\subsection{Learning Destroy Operator}
\label{subsec:learning}

We present the learning based destroy operator for LNS to automatically select the variables. To this end, we first formulate the LNS as a Markov Decision Process (MDP), and then parameterize the policy by a bipartite GCN since the MILP model could be represented as a graph (an illustration of our network architecture is depicted in Fig. \ref{fig_2}), which is trained by the imitation learning algorithm.

\subsubsection{MDP Formulation}
For an instance of MILP model $p$ with integer decision variables $\mathcal{X}$, the solving process by LNS can be represented as a MDP. Specifically, we consider the destroy operator as the \emph{agent} and the remainder of LNS as the \emph{environment}. At the $t^{th}$ iteration, the \emph{state} $s_t$ consists of both the formulation information and the current solution. The \emph{action} $a_t$ is to select a subset of variables among all integer decision variables $\mathcal{X}$ according to the \emph{policy} $\pi(a_t|s_t)$. After the action is taken, the environment returns a new solution by reoptimizing the selected variables via the repair operator, and also updates state to $s_{t+1}$. The reward is represented as the difference between the objective values of the new and current solution, i.e., $r(s_t, a_t) = c(x_{t+1})-c(x_t)$, where $x_t$ and $x_{t+1}$ are reflected in states $s_t$ and $s_{t+1}$. This interaction between the agent and environment continues until the pre-defined termination condition is met. 

\subsubsection{State Representation}
\label{sub_feature}
We represent the state as a bipartite graph $\mathcal{G}=(\mathcal{C},\mathcal{V},\mathcal{E})$, with constraints $c_i\in \mathcal{C}$, variables $v_j\in \mathcal{V}$ and edges $e_{ij}=(c_i,v_j)\in \mathcal{E}$, which is depicted in Fig. \ref{fig_2}(a). The red nodes $\mathcal{C}$ refer to constraints, i.e., rows in the adjacency matrix of the MILP formulation. They are represented by a feature matrix $\mathcal{M}_c\in \mathbb{R}^{|\mathcal{C}|\times n_c}$, where $n_c$ is the dimension of constraint features. The blue nodes $\mathcal{V}$ refer to variables, i.e., columns in the adjacency matrix of the MILP formulation. They are represented by a feature matrix $\mathcal{M}_v\in \mathbb{R}^{|\mathcal{V}|\times n_v}$, where $n_v$ is the dimension of variable features. The edges $\mathcal{E}$ refer to connections between variables and constraints (i.e., an edge $e_{ij}$ exists if the coefficient of variable $v_j$ is nonzero in constraint $c_i$). They are represented by a feature matrix $\mathcal{M}_e\in \mathbb{R}^{|\mathcal{C}|\times|\mathcal{V}|\times n_e}$, where $n_e$ is the dimension of edge features. We represent the features for each constrain $c_i$, variable $v_j$, edge $e_{ij}$ as $\mathbf{c}_i^0$, $\mathbf{v}_j^0$, $\mathbf{e}_{ij}^0$, respectively, and describe them in TABLE \ref{table_2}, where $\mathbf{w}_h^0$ refers to the feature of each vehicle $w_h$ (denoted by the green nodes in Fig. \ref{fig_2}(c)). 

\begin{table}[!t]
  \newcommand{\tabincell}[2]{\begin{tabular}{@{}#1@{}}#2\end{tabular}}
  \setlength{\tabcolsep}{3.7pt}  
  \renewcommand{\arraystretch}{1.2}
  \caption{Features for constraint, edge, variable and vehicle (Dim represents the dimension of each feature).}
  \label{table_2}
  \centering
  \footnotesize
  \begin{tabular}{c|l|l|c}
  \toprule
   & Features & Description & Dim\\
  \hline
  \multirow{3}*{$\mathbf{c}_i^0$} & obj\_cos\_sim & Cosine similarity with objective. & 1\\
  \cline{2-4}
  & bias & Normalized bias value. & 1\\
  \cline{2-4}
  & type & The constraint form (=, $\leq$). & 2\\
  \hline
  $\mathbf{e}_{ij}^0$ & coef & Normalized constraint coefficient. & 1\\
  \hline
  \multirow{4}*{$\mathbf{v}_j^0$} & type & Type (Integer, continuous). & 2\\
  \cline{2-4}
  & coef & Normalized objective coefficient. & 1\\
  \cline{2-4}
  & sol\_val & Solution value. & 1\\
  \cline{2-4}
  & inc\_val & Value in incumbent. & 1\\
  \hline
  \multirow{2}*{$\mathbf{w}_h^0$} & is\_used & Whether used or not in a solution. & 2\\
  \cline{2-4}
  & sol\_cost & Normalized cost (distance) in a solution. & 1\\
  \bottomrule
  \end{tabular}
\end{table}

\subsubsection{Policy Network}
We parameterize the policy by a graph convolutional network (GCN) $\pi_\theta$, given its desirable capability in learning representation over graph-structured inputs and generalizing to different sizes. Specifically, given the state representation in Fig. \ref{fig_2}(a), the network first executes the graph convolution by applying two sequential message passing, i.e., from variable to constraint nodes ($\mathcal{V}$ to $\mathcal{C}$), and from constraint to variable nodes ($\mathcal{C}$ to $\mathcal{V}$), respectively. Such convolution process is expressed as follows,
\begin{gather}
  \mathbf{c}_i \gets f_1([\mathbf{c}_i; g_1(\sum_{v_j}^{e_{ij}\in \mathcal{E}} \mathbf{e}_{ij}^0\mathbf{v}_j)]),\,\, \forall c_i\in \mathcal{C} \\
  \mathbf{v}_j \gets f_2([\mathbf{v}_j; g_2(\sum_{c_i}^{e_{ij}\in \mathcal{E}} \mathbf{e}_{ij}^0\mathbf{c}_i)]),\,\, \forall v_j\in \mathcal{V}
\end{gather}
where $\mathbf{c}_i$ and $\mathbf{v}_j$ represent the embeddings of constraint $c_i$ and variable $v_j$, which are initialized by linearly projecting their raw features $\mathbf{c}_i^0$ and $\mathbf{v}_j^0$ into 64-dimensional feature spaces, respectively; $[ ; ]$ means the concatenation operation; $f_1$, $f_2$, $g_1$, and $g_2$ are multilayer perceptrons (MLPs) with a hidden layer and a ReLU activation. We keep all MLPs having hidden dimensions 128 and output dimensions 64, so that all embeddings are 64-dimensional. Accordingly, the variable and constraint embeddings are updated by the above convolutions, as shown in Fig. \ref{fig_2}(b).

Since we aim to select partial variables for reoptimization, we aggregate the variable embeddings by \textit{mean pooling} to nodes $\mathcal{W}$, each of which corresponds to a group of variables with the same index of fleet and vehicle. Then, the aggregated embeddings over each group of variables $w_h\in \mathcal{W}$ are concatenated with the 64-dimensional linear projection of the feature matrix $\mathcal{M}_w\in \mathbb{R}^{|\mathcal{W}|\times n_w}$ of nodes $\mathcal{W}$. The raw features of each vehicle in fleets ($\mathbf{w}_h^0$) are contained in each row of $\mathcal{M}_w$ and described in TABLE \ref{table_2}.
The above aggregation process (in Fig. \ref{fig_2}(c)) is expressed as follows,
\begin{gather}
  \mathbf{w}_h \gets [\frac{1}{|\eta(w_h)|}\sum_{v_j\in \eta(w_h)} \mathbf{v}_j; \mathbf{w}_h],\,\, \forall w_h\in \mathcal{W}
\end{gather}
where $\mathbf{w}_h$ denotes the embedding of $w_h$ and is initialized by linear projection of its raw features $\mathbf{w}_h^0$; $\eta(\cdot)$ is a function that gathers the set of variables with respect to a node $w_h$. Thus, the embedding $\mathbf{w}_h$ represents a group of variables related to the same vehicle in a problem instance. As depicted in Fig. \ref{fig_2}(d), we finally process these embeddings by another MLP ($f_3$) that is similar to the previous ones but with a single output, followed by sigmoid activation, such that,
\begin{gather}
  P(w_h)= \text{Sigmoid}(f_3(\mathbf{w}_h)),\,\, \forall w_h\in \mathcal{W}.
\end{gather}

Consequently, we attain the probability for each group of variables $w_h\in \mathcal{W}$, with which we could further sample variables related to certain vehicles for reoptimization.

\begin{algorithm}[!t]
  \caption{Forward Training for LNS}
  \label{FT}
	\begin{algorithmic}[1]
		\STATE Input: a set of MILP instances $\{p_i\}_{i=1}^m$ with initial solutions $\{x_i\}_{i=1}^m$; number of iterations $n_i$; number of sampling for an instance $n_s$; Vehicle-random destroy operator $\mathcal{D}$; repair operator $\mathcal{R}$; percentage of decision variables to destroy $\mathcal{D}_d$.
		\FOR{$t=1,\cdots,n_i$}
		    \STATE $D_t = []$
    		\FOR{$i=1,\cdots,m$}
    		    \STATE $best\_obj, best\_D = c(x_i),$ None
    		    \FOR{$j=1,\cdots,n_s$}
    		        \STATE $\_, x_j, \mathcal{X}_r$ = \textbf{Algorithm 1}$(p_i, x_i, 1, \mathcal{D}, \mathcal{R}, \mathcal{D}_d$)
    		        \IF{$c(x_j) < best\_obj$}
    		        \STATE $best\_D = \mathcal{X}_r$
    		        \STATE $best\_obj = c(x_j)$
    		        \ENDIF
    		    \ENDFOR
    		    \IF{$best\_D$ is not None}
        		    \STATE $D_t.append((p_i, x_i, best\_D))$
        		\ENDIF
    		\ENDFOR
    	    \STATE $\pi_t$ = SUPERVISE\_TRAIN$(\pi_{t-1}, D_t)$
    	    \FOR{$i=1,\cdots,m$}
    	    \STATE $\_, x_i, \_$ = \textbf{Algorithm 1}$(p_i, x_i, 1, \pi_t, \mathcal{R}, \mathcal{D}_d)$
    	    \ENDFOR
		\ENDFOR
		\RETURN $\pi_1, \pi_2, \cdots, \pi_{n_i}$
	\end{algorithmic}
\end{algorithm}

\subsubsection{Imitation Learning}
We train our policy by imitation learning to mimic desirable behaviors of a heuristic destroy operator, which avoids learning from scratch in the huge combinatorial search space. Here, we apply our \emph{Vehicle-random} multiple times and retrieve the best one to serve as demos for imitation learning. The well-known behavior \emph{cloning} is a typical imitation learning algorithm \cite{pomerleau1998autonomous}, but it suffers from quadratic-growth compounding errors along the time horizon. In our method, we exploit a \emph{forward training} algorithm, which is able to correct the cascading errors \cite{ross2010efficient} by adaptively collecting demos based on the learned policy. The training details in one epoch are displayed in Algorithm \ref{FT}. Particularly, at the $t^{th}$ LNS iteration, we collect demos $D_t$ by running \emph{Vehicle-random} destroy operator $n_s$ times on each training instance, and retrieving the ones with the best objectives (Lines 3-16). The demos with non-improved solutions $x_j$ on some instances will not be collected (Line 13). Accordingly, $D_t=\{(p_u, x_u, a_u)\}_{u=1}^{\mathcal{N}_t}$ with $\mathcal{N}_t$ denoting the number of collected demos in $t^{th}$ iteration. Each element in $D_t$ is a 3-tuple that contains the problem instance, current solution and action (i.e., the selected variables). We then represent them by the respective state as described in section \ref{sub_feature}, and thus obtain $D_t=\{(s_u,a_u)\}_{u=1}^{\mathcal{N}_t}$. Thereafter, the policy network is trained with the collected data in a supervised manner (Line 17), to minimize the binary cross-entropy loss as follows,
\begin{equation}
\begin{aligned}
\mathcal{L(\theta)}=-\frac{1}{\mathcal{N}_t|\mathcal{W}|}&\sum_{(s,a)\in D_t}\sum_{w_h\in \mathcal{W}}[y(w_h)\cdot\text{log}P(w_h)\\
&+(1-y(w_h))\cdot\text{log}(1-P(w_h))],
\end{aligned}
\end{equation}
where $y(w_h)=1$ if the group of variables $w_h$ is selected otherwise $y(w_h)=0$. In other words, we regard the action of selecting a subset of variables as determining whether a group of variables related to the same vehicle is selected. Given a state, it comes down to a binary classification problem for each group of variables. After the policy network is updated with the demos in iteration $t$, the new solution to each instance is attained by running the LNS with the trained policy $\pi_t$ (Lines 18-20), and we continue collecting new data in the next iteration to repeat the training.


\subsubsection{Deployment}
\label{subsec:utilization}
Given the trained policy, we could apply it in the LNS framework as depicted in Fig. \ref{fig_1}. Regarding how to select the groups of variables according to the output probabilities, we could deploy it in three different ways, i.e., \textit{fix-sized sampling}, \textit{adaptive sampling} and \textit{disjoint sampling}, which accordingly yields three LNS variants as follows.

\textbf{IL-sample:} Given the destroy degree, we sample a fix-sized subset of variables in each LNS iteration, based on a softmax over probabilities for each group of variables $w_h$, such that, $\sigma_{w_h}=e^{P(w_h)}/{\sum}_{\kappa=1}^{|\mathcal{W}|} e^{P(w_\kappa)}, \,\, \forall w_h\in \mathcal{W}$. This deployment selects variables with a high potential to improve the solution.

\textbf{IL-sampleA:} Each group of variables is sampled independently based on its output probability $P(w_h), \forall w_h\in \mathcal{W}$, such that the number of variables to be optimized in each iteration is stochastic. 

\textbf{IL-sampleD:} It follows the way in IL-sample, but imposes that the sampled subsets between consecutive LNS iterations are disjoint.

\section{Experiments}
\label{sec:experiments}
In this section, we evaluate our method for solving the vehicle routing problem in AGH with more realistic scenarios, where instances of 20, 50, 100 flights and 10 types of operations are first considered. We then compare our learning based LNS with the state-of-the-art commercial solver CPLEX (i.e., also the repair operator in our LNS) and conventional LNS heuristics to verify its superiority. We also show that our learning based LNS is able to generalize well to large instances even if the policy is trained on small instances. Subsequently, we verify the versatility of our method by adopting a different solver, i.e., OR-Tools, as the repair operator. Finally, we further evaluate the scalability of our method on larger instances of up to 200 flights. The code is publicly available at: \url{https://github.com/RoyalSkye/AGH}.

\begin{figure}[!t]
\centering
\includegraphics[scale=0.24]{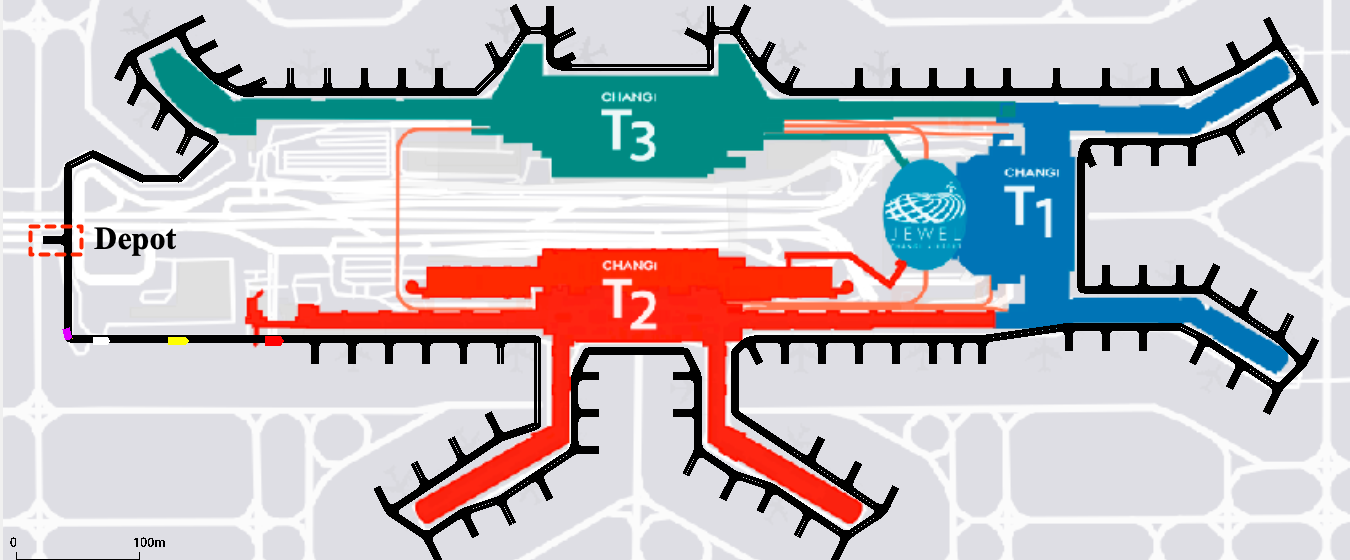}
\caption{The map of CHANGI airport with 3 terminals and 90 gates (where the most left point is the depot).}
\label{airport}
\vspace{-2mm}
\end{figure}

\subsection{Settings}
\noindent\textbf{Instance Generation.} Our experiments are conducted on a well-known international airport in Singapore, i.e.,  CHANGI airport, whose topology is shown in Fig.~\ref{airport}. It contains 3 terminals with 90 gates where the operations are executed. We load the map\footnote{\url{https://www.changiairport.com/en/maps.html}} into SUMO~\cite{SUMO2018}, by which the distances between gates and the depot are calculated automatically. On top of the map, we first generate AGH instances of 20, 50 and 100 flights which are referred as AGH20, AGH50 and AGH100, respectively. Particularly, the instances are generated based on statistics of the real scenario in CHANGI airport. For the flight density, we follow the scheduling table in one day\footnote{\url{https://www.changiairport.com/en/flights.html}}, and randomly generate the number of hourly aircraft arrivals from [5, 25], where 5 and 25 correspond to the flight quantities in the idlest and busiest periods, respectively. The duration of a turnaround for each aircraft is sampled from [30, 60] minutes. We consider three types of aircraft with respective precedence relations according to their manufacturers \cite{airbus1995a320,airplanes2005767}, the first one of which has been shown in Fig. \ref{fig_0}. The others are similar to the first type yet much harder, i.e., the priority of catering could be 1 or 2 for the second type; the cleaning could be prioritized as 1 or 2 and catering can be executed anytime before pushback for the third type. The three aircraft types are uniformly across all the flights, which are also uniformly assigned to the gates. For the 10 operations to be scheduled, their service durations, time windows and vehicle speeds are set individually following \cite{changi,de2010ground}. The number of vehicles in each fleet is randomly sampled from [10, 20]. We randomly generate the flight demands for each operation from [5, 15], and set the ratio of demand to capacity to a random value in [0.7, 0.9] following \cite{chao1999computational}.

\begin{table*}[!t]  
\setlength{\tabcolsep}{1.2pt}
\renewcommand{\arraystretch}{1.3}
	\centering
	\caption{Comparison with Baselines.}
	\begin{tabular}{l cccrrc cccrrc cccrrc}
		\toprule
	    & \multicolumn{6}{c}{AGH20 (6k/15k)} & \multicolumn{6}{c}{AGH50 (30k/140k)}   & \multicolumn{6}{c}{AGH100 (120k/700k)} \\
		\cmidrule(lr){2-7}\cmidrule(lr){8-13}\cmidrule(lr){14-19}
			Methods & Mipgap & Limit & $\mathcal{D}_d$ & Obj. & Gap & Time & Mipgap & Limit & $\mathcal{D}_d$ & Obj. & Gap & Time & Mipgap & Limit & $\mathcal{D}_d$ & Obj. & Gap & Time \\ 
			\midrule
			CPLEX   & 1e-04 & 1m & -- & 86384 & 12.61\% & 1m & 1e-04 & 5m & -- & 365353 & 152.80\% & 5m & 1e-04 & 30m & -- & 696762 & 178.58\% & 30m \\
			CPLEX ($\times 5$)   & 1e-04 & 5m & -- & \textbf{77630} & \textbf{1.65\%} & 5m & 1e-04 & 25m & -- & 341397 & 134.65\% & 25m & 1e-04 & 2.5h & -- & 684509 & 173.82\% & 2.5h \\
			Decomposition~\cite{padron2019improved}   & 1e-04 & 45m & -- & 107269 & 41.44\% & 45m & 1e-04 & 47m & -- & 215620 & 47.70\% & 47m & 1e-04 & 57m & -- & 369966 & 48.01\% & 57m \\
			GA~\cite{liu2021scheduling}   & -- & 1h & -- & 115706 & 51.71\% & 1h & -- & 1h & -- & 261137 & 77.60\% & 1h & -- & 1h & -- & 438542 & 76.74\% & 1h \\
			Random   & 0.1 & 10s & 0.4 & 151562 & 99.37\% & 1m & 0.1 & 1m & 0.4 & 332537 & 128.76\% & 5m & 0.2 & 10m & 0.4 & 491285 & 96.42\% & 30m \\
			Vehicle-random   & 0.1 & 10s & 0.4 & 86301 & 13.04\% & 1m & 0.1 & 1m & 0.4 & 166735 & 14.66\% & 5m & 0.2 & 10m & 0.4 & 311035 & 24.21\% & 30m \\
			Vehicle-worst   & 0.1 & 10s & 0.4 & 98383 & 28.31\% & 1m & 0.1 & 1m & 0.4 & 223397 & 53.43\% & 5m & 0.2 & 10m & 0.4 & 663911 & 164.62\% & 30m \\
			Vehicle-worst-random   & 0.1 & 10s & 0.4 & 90477 & 18.32\% & 1m & 0.1 & 1m & 0.4 & 170875 & 17.58\% & 5m & 0.2 & 10m & 0.4 & 316498 & 26.31\% & 30m \\
			IL-sample (Ours)   & 0.1 & 10s & 0.4 & 85942 & 12.53\% & 1m & 0.1 & 1m & 0.4 & \textbf{166080} & \textbf{14.24\%} & 5m & 0.2 & 10m & 0.4 & \textbf{302512} & \textbf{20.86\%} & 30m \\
		\bottomrule          
	\end{tabular}
	\label{table_3}
\end{table*}

\noindent\textbf{Baselines.} We compare our learning based LNS with the state-of-the-art exact solver for MILP, and various conventional LNS heuristics. In specific, the baselines include, 1) the exact solver CPLEX~\cite{cplex2009v12} (we also use OR-Tools~\cite{ortools} instead to verify the versatility of our method), which is also the repair operator in the LNS methods; 2) \textit{Random}, which destroys the current solution by randomly picking a subset of integer decision variables to reoptimize; 3) \textit{Vehicle-random}, which is used to collect demos for imitation learning as described in section~\ref{subsec:destroy}; 4) \textit{Vehicle-worst}, which sorts all used vehicles according to their travelling distance in the current solution, and selects the variables related to those vehicles with the longest distances; 5) \textit{Vehicle-worst-random}, which selects $\mathcal{D}_d/2$ of variables with Vehicle-random and $\mathcal{D}_d/2$ of variables with Vehicle-worst, given a destroy degree $\mathcal{D}_d$; 6) \textit{Decomposition}, which tackled a similar problem to ours and reported favorable performance by first decomposing it into sub-problems for each fleet using constraint programming (CP) and then applying LNS heuristic to solve them~\cite{padron2019improved}. We adapted it to solve the same problem as ours. Note that the work in \cite{padron2019improved} is the recent one for solving AGH problems whose settings are most similar to ours \cite{ng2018review}; 7) \textit{GA}, which solved the AGH problem by a genetic algorithm~\cite{liu2021scheduling}. While its original work only considers one type of aircraft, we adapt the GA method to our setting with heterogeneous aircraft. Other specialized approaches for VRPs with individual operations are not considered as baselines, since they are less relevant to the studied problem in this paper.


For LNS heuristics and our method, we generate initial solutions with a simple heuristic based on nearest neighbor insertion~\cite{solomon1987algorithms}, which is commonly used to initialize the solutions for routing problems~\cite{battarra2009adaptive,yin2013mathematical,braysy2004evolutionary}. Specifically, each route, starting from the depot, is iteratively extended with the nearest unserved flight (which also satisfies the basic constraints), until no such flights exist and then a new route starts. Meanwhile, the routes are constructed for each fleet following the precedence relation among each other. We note that the above insertion heuristic basically picks the nearest aircraft which satisfies all constraints and it scarcely needs domain knowledge for the problem. Alternatively, we can also simply run a solver (e.g. CPLEX) with a time limit to attain initial solutions.


\noindent\textbf{Hyperparameters in LNS Heuristics.} We use CPLEX 12.10 as the solver in baseline 1, and as the repair operators in baselines 2-5 and our method.  We set two parameters in CPLEX (i.e., mipgap\footnote{The relative gap between the lower and upper objective bound.} and timelimit) as stopping conditions in each LNS iteration. The current LNS iteration stops when either of the two conditions is met. In particular, we set mipgap to 0.1 for AGH20 and AGH50, 0.2 for AGH100. The timelimit is set to 10s, 1m and 10m respectively, though they could be longer than the real runtime in iterations. The acceptance criterion is that in each iteration, we only accept the solution whose objective value is smaller or at most $1\%$ larger than that of the incumbent. Typically, we find that all LNS heuristics considerably improve solutions between 5 and 10 iterations, so we empirically set the total runtime to 1m, 5m and 30m for AGH20, AGH50 and AGH100, respectively. Since the destroy degree cannot be too large or small (as analyzed in Section 4.2), we have tried 0.3, 0.4, 0.5 and 0.6 in our experiment. We observe that when it equals to 0.4, all LNS heuristics (including Vehicle-random LNS that generates demos for the policy network) perform relatively well on the validation set. Therefore, we set it to 0.4 for all LNS heuristics on AGH20, AGH50 and AGH100.


\noindent\textbf{Hyperparameters in Learning based LNS.} We train the model only on instances of AGH20 to learn the policy (which consumes about five days), and test it on instances of AGH20, AGH50 and AGH100 to evaluate its performance. Regarding AGH20, we generate 70 instances for training, 30 instances for validation and 50 instances for testing. For each of other sizes, we also generate 50 instances for testing. During the training, we set the epoch number to 10 according to the convergence of the training curve. In each epoch, we set the iteration number of LNS $n_i=5$, since we observe its performance cannot be significantly improved afterwards. In each iteration of the forward training algorithm (i.e., each LNS iteration), we run Vehicle-random LNS $n_s=10$ times for each instance in the training set, and keep the best one as a demo. Accordingly, the stopping condition, acceptance criterion and destroy degree are set following the hyperparameters of Vehicle-random LNS, as mentioned above. After the demos are collected, we train the neural network by the stochastic gradient descent (SGD) optimizer \cite{ruder2016overview}, with the learning rate equals to $10^{-4}$ and the batch size equals to 16 due to GPU memory limit. For validation, we set the iteration number $n_i=10$ and the destroy degree $\mathcal{D}_d=0.4$. For testing, we keep the destroy degree same as the training, and run the learned LNS with the same time as other baselines.
All experiments are conducted on a machine with 32 AMD EPYC 7601 32-Core Processors. 


\begin{figure*}[!ht]
  \centering
  \subfloat[AGH20]{\includegraphics[width=2.3in]{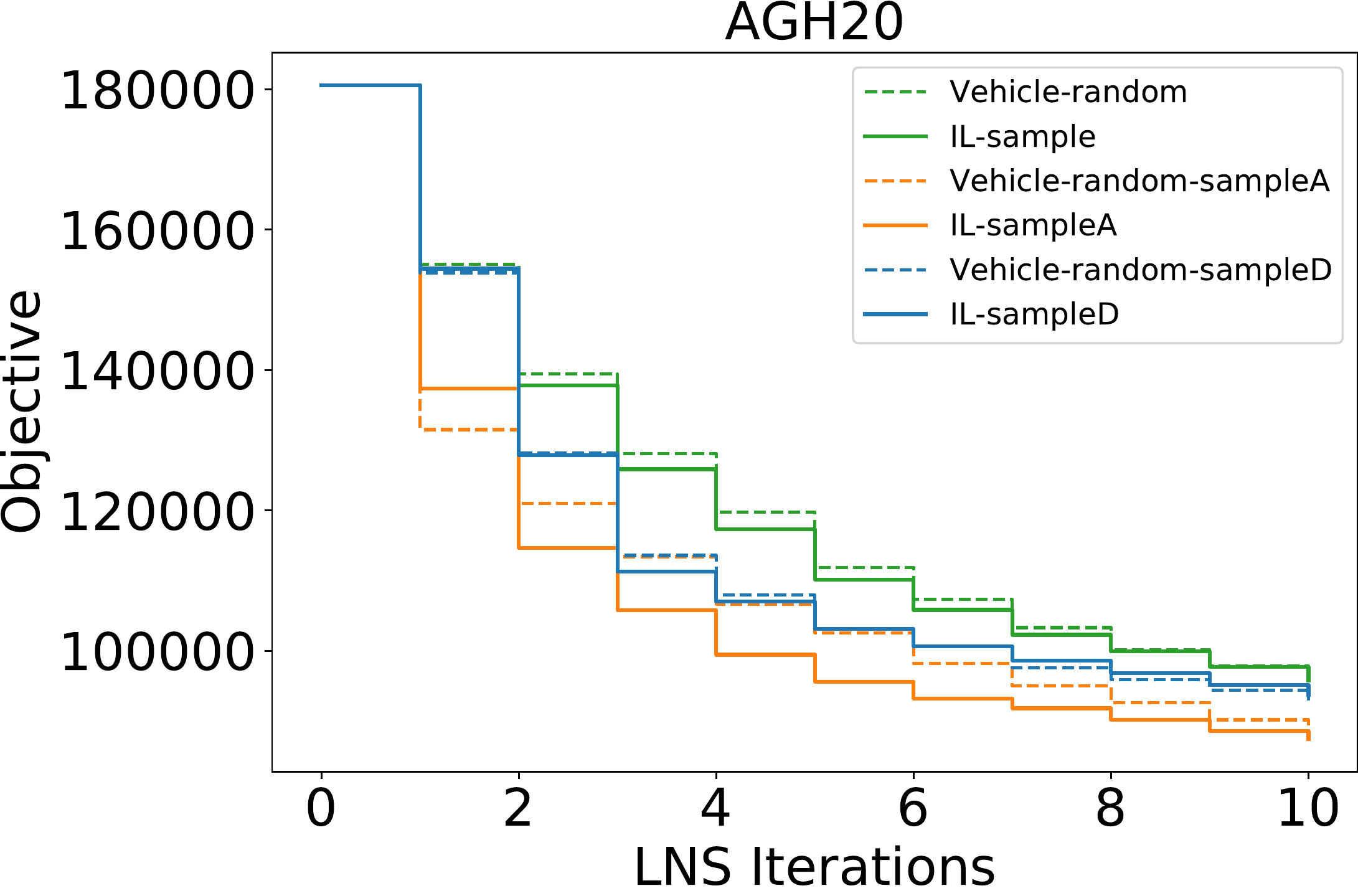}%
  \label{fig_4_1}}
  \hfil
  \subfloat[AGH50]{\includegraphics[width=2.3in]{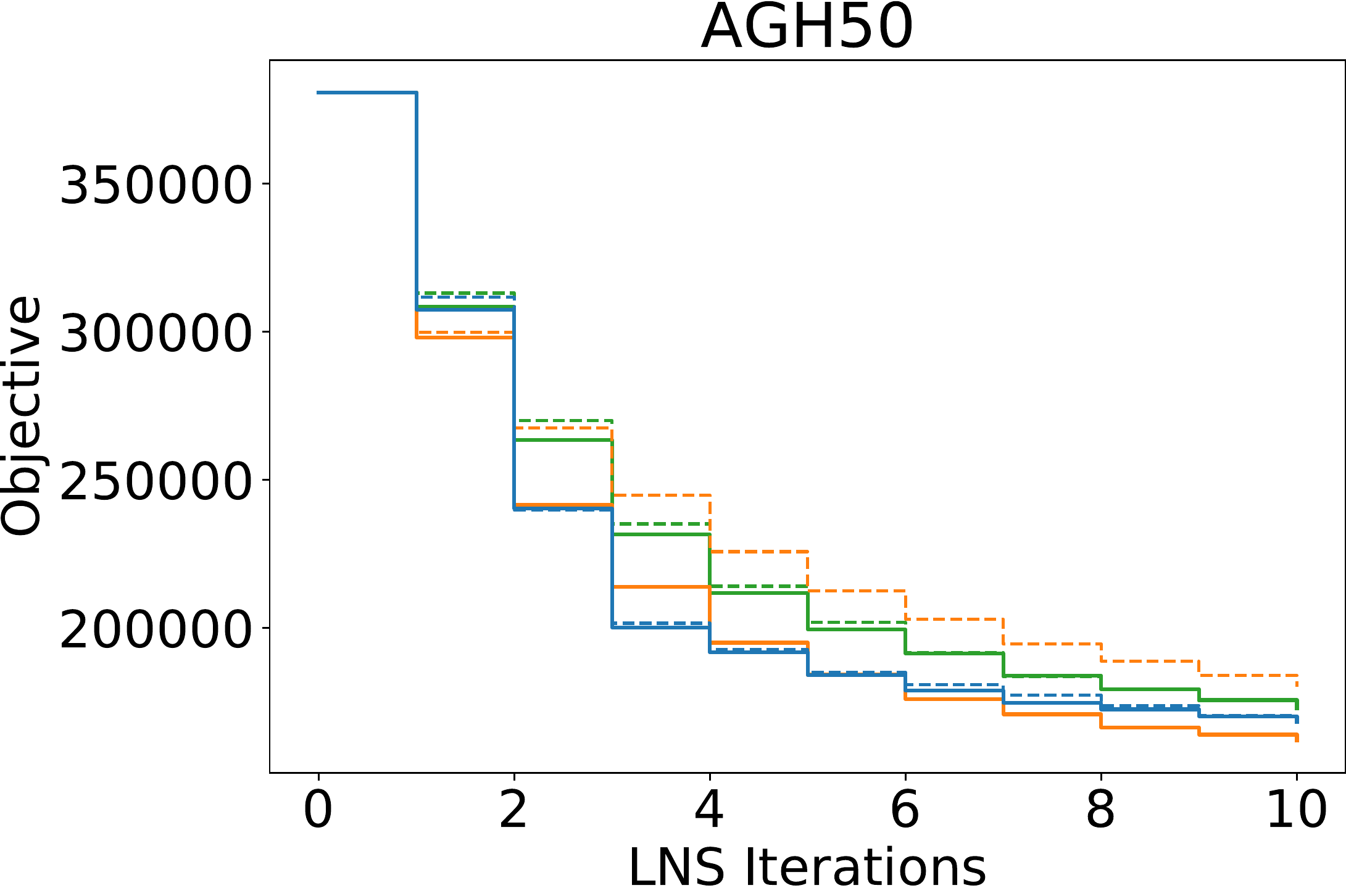}%
  \label{fig_4_2}}
  \hfil
  \subfloat[AGH100]{\includegraphics[width=2.3in]{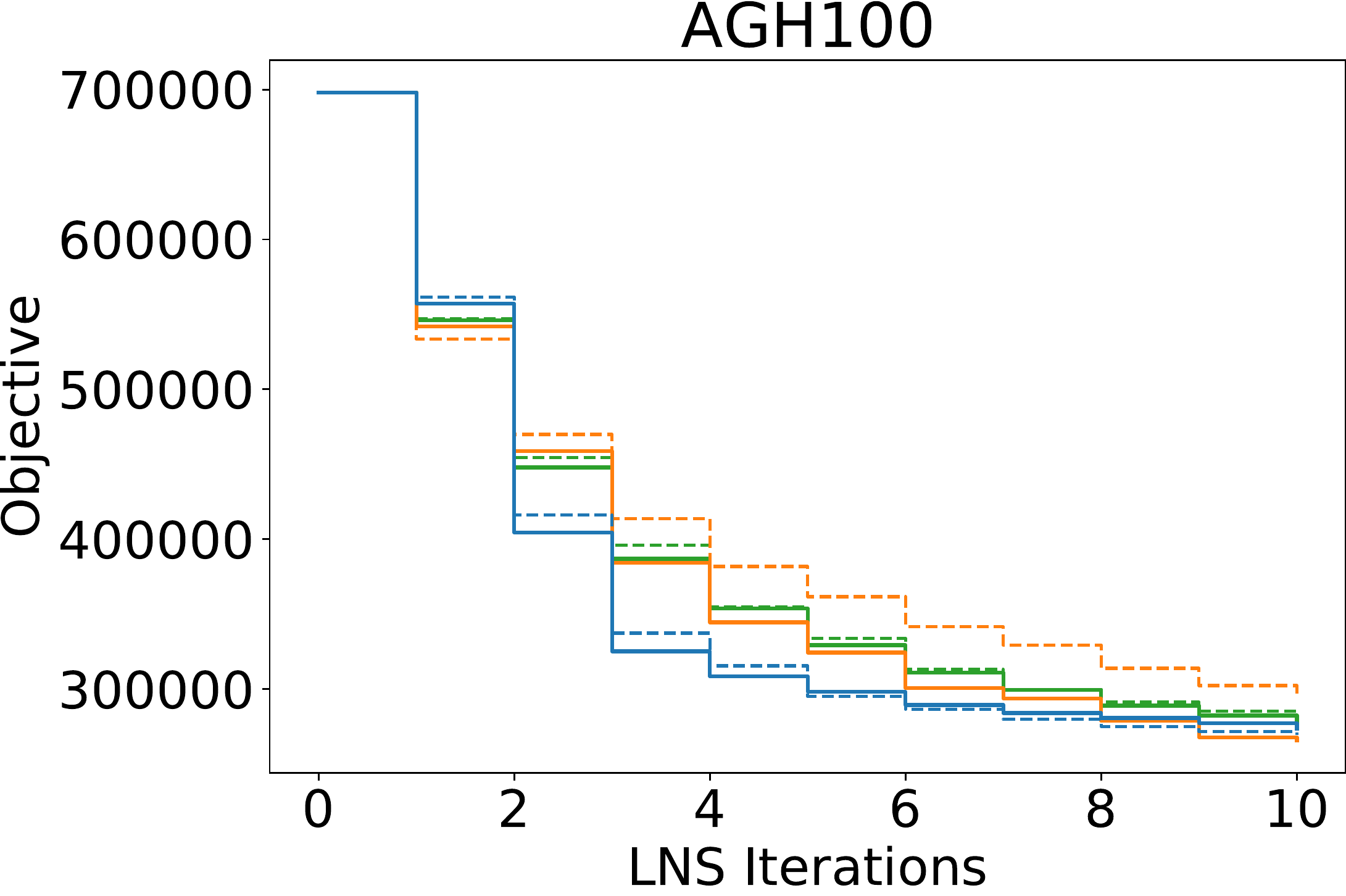}%
  \label{fig_4_3}}
  \caption{The objective values of incumbents against iterations for the learned LNS and heuristic counterparts.}
  \label{fig_4}
  \vspace{-2mm}
\end{figure*}
\begin{figure*}[!ht]
  \centering
  \subfloat[AGH20]{\includegraphics[width=2.3in]{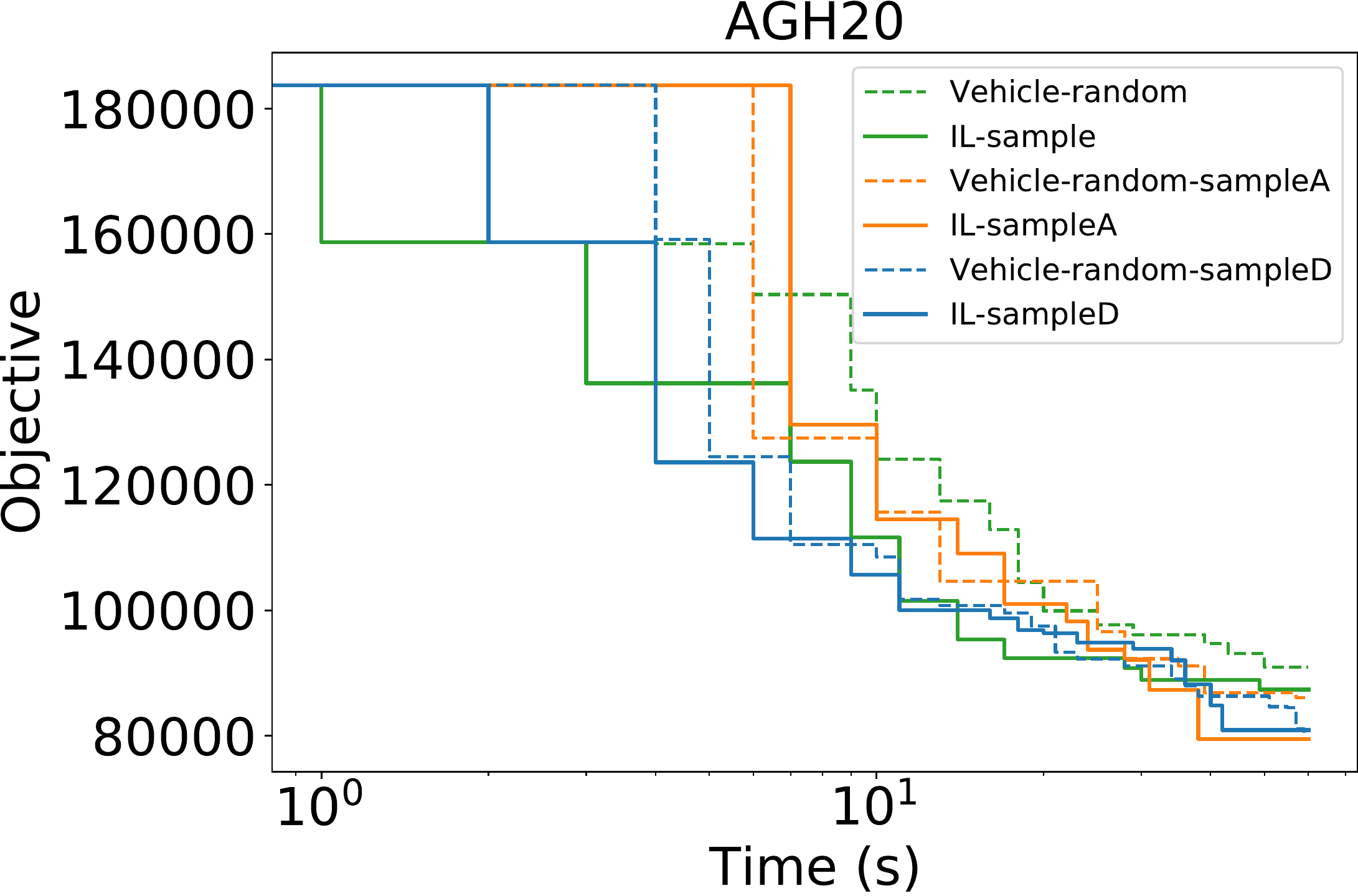}%
  \label{fig_5_1}}
  \hfil
  \subfloat[AGH50]{\includegraphics[width=2.3in]{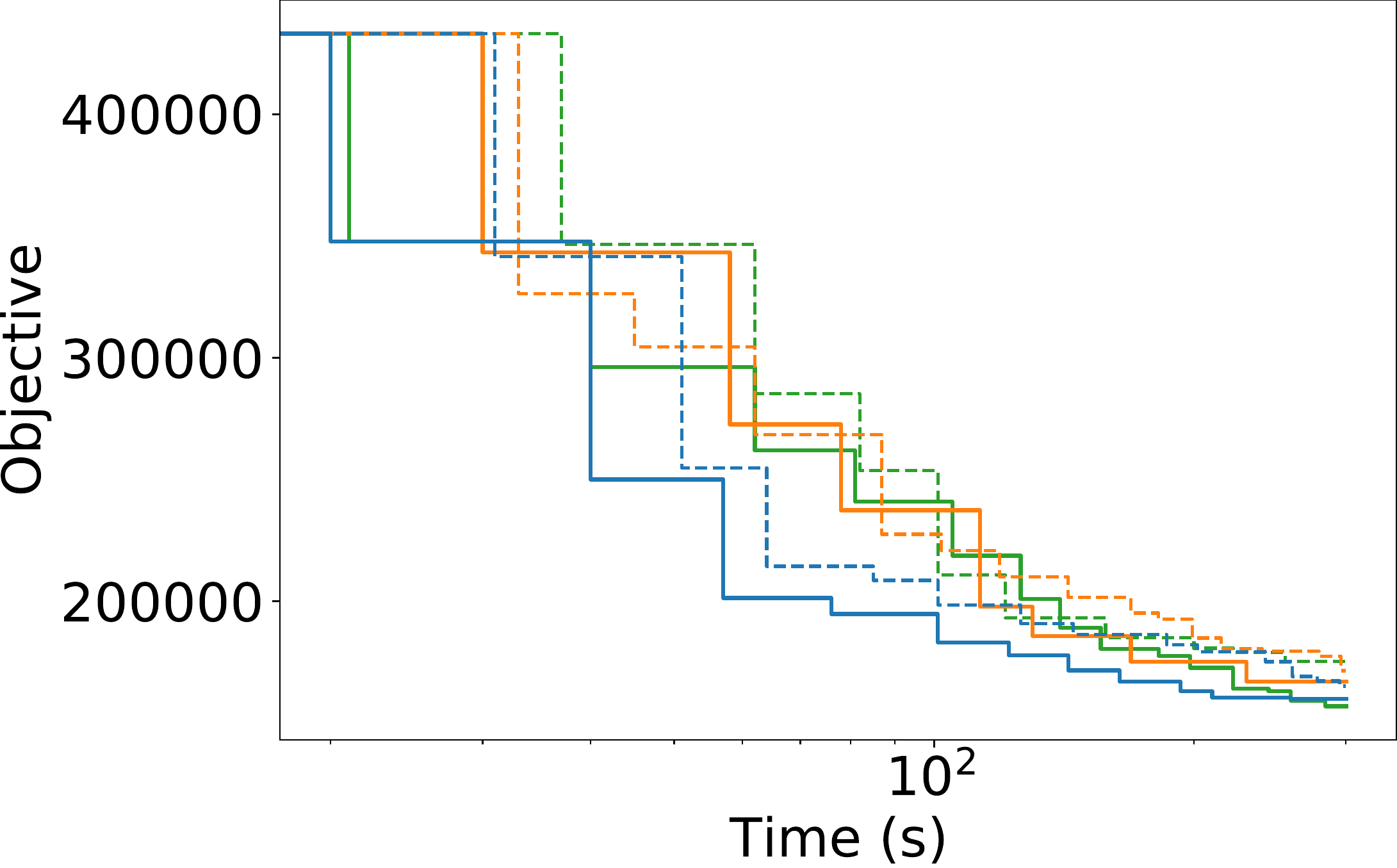}%
  \label{fig_5_2}}
  \hfil
  \subfloat[AGH100]{\includegraphics[width=2.3in]{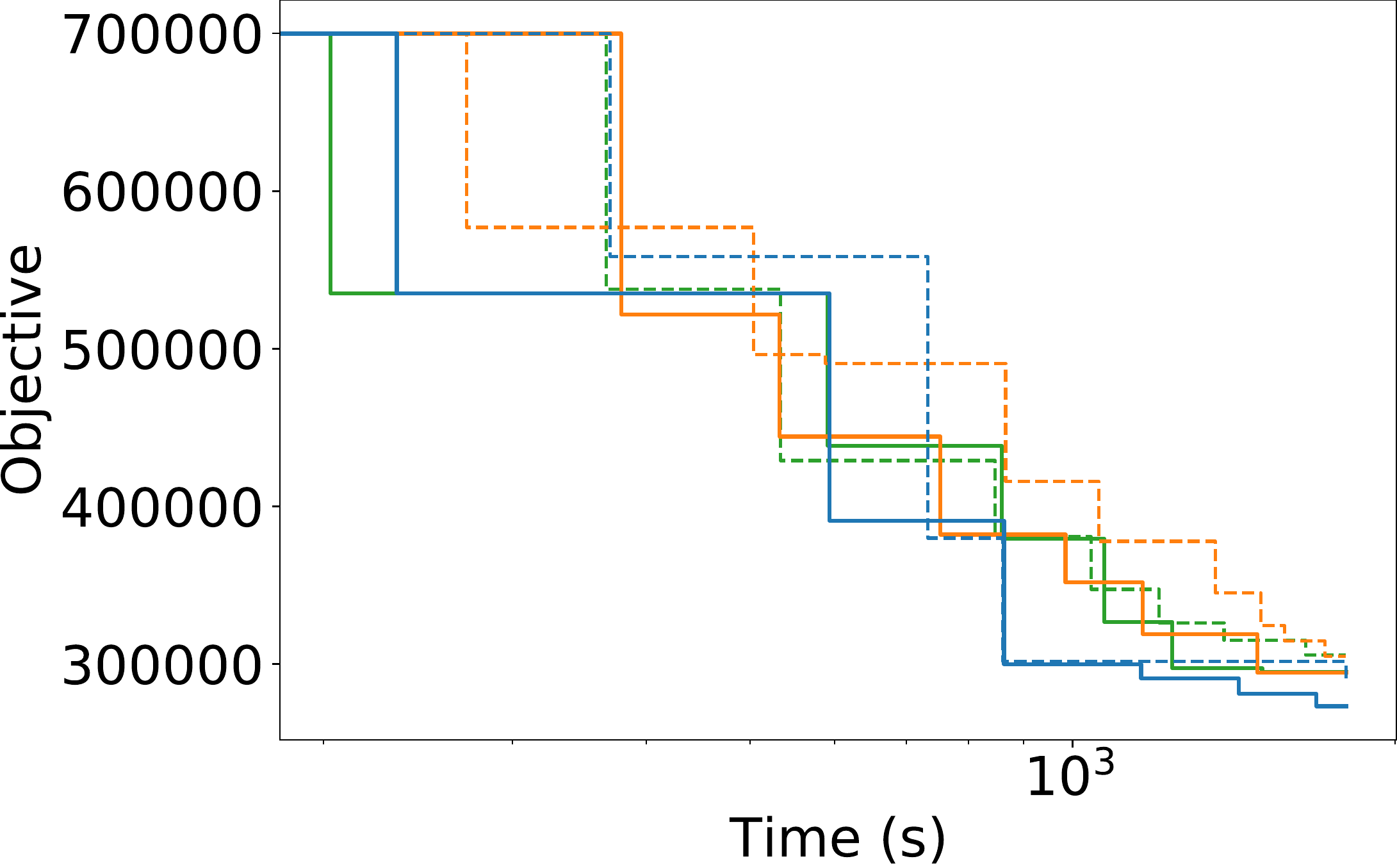}%
  \label{fig_5_3}}
  \caption{The objective values of incumbents against runtime for the learned LNS and heuristic counterparts.}
  \label{fig_5}
  \vspace{-2mm}
\end{figure*}

\subsection{Comparison Study}
\label{exp-2}


We compare the learning based LNS with other baselines including CPLEX solver, conventional LNS heuristics (with CPLEX as the repair operator), the decomposition method in~\cite{padron2019improved} and the GA method adapted from~\cite{liu2021scheduling}. We set the total runtime of the above methods to 1m, 5m, 30m for AGH20, AGH50 and AGH100, respectively, since beyond which we found that solutions in LNS methods only have marginal improvement. Regarding CPLEX, we not only apply it to solve the instances of the MILP model using the same time with other methods, but also allow it to run as 5 times long as others. For the method in~\cite{padron2019improved}, we reproduce the algorithm using its original parameters, and allow it to run much longer time than ours. We also allow the GA method to run a long time, i.e., 1h for all problem scales. For a fair comparison, we only deploy the learned destroy operator with the fix-sized sampling strategy as stated in section \ref{subsec:utilization}, i.e., IL-sample, and keep $\mathcal{D}_d=0.4$.
Then we compare the average objective value of the best found solutions over the testing set for all methods, and the primal gap\footnote{We first compute 
$|c^{\top}\overline{x}-c^{\top}\overline{x}^{\ast}|/max\{|c^{\top}\overline{x}|,|c^{\top}\overline{x}^{\ast}|\}\cdot100\%$ for each instance, where $\overline{x}$ is the solution by a method and $\overline{x}^{\ast}$ is the best one out of all methods, and then average the gaps over the testing set.}\cite{khalil2017learn} since we are unable to acquire the optimal solution to calculate the optimality gap. All results are gathered in Table \ref{table_3}, where we also display the average number of constraints and variables (in brackets) over the MILP formulations for different problem sizes, along with some key parameters for each method. We can easily observe that, 1) all LNS methods including our learning based one, with the same runtime, can significantly outperform the CPLEX solver on AGH50 and AGH100 in terms of objective value and gap; 2) though CPLEX has an obvious advantage over other LNS methods on AGH20, the learned LNS by our method still outstrips it with the same runtime; 3) when running with much longer time, CPLEX can achieve the best result on AGH20 since the instances are relatively small and tractable for the exact solver, while it is inferior to LNS methods on AGH50 and AGH100, with a growing gap as the problem scales up; 4) for LNS methods, random destroy operator has clearly larger gap than the three vehicle based destroy operators, where Vehicle-random and Vehicle-worst-random demonstrate similar performance and outperform Vehicle-worst; 5) the learned LNS surpasses all other LNS heuristics on AGH20, AGH50 and AGH100, indicating that our method is able to learn a more effective destroy operator to guide the LNS algorithm; 6) despite the extra runtime, the decomposition and GA method are generally comparable to some LNS heuristics but far inferior to our learnt one. 


The method in~\cite{padron2019improved} adopts constraint programming (CP) to decompose the original problem into a VRPTW for each fleet, then applies LNS to solve each resulting VRPTW. In the LNS, the random pivot and small routing operators~\cite{guimarans2012hybrid,rousseau2002using} are particularly designed to remove flights or edges in the route based on various criteria (e.g., geographic and temporal proximity), and CPLEX solver is used to rebuild the destroyed solution. In contrast, we exploit learned LNS to decompose the original problem by directly selecting variables from the MILP formulation, which circumvents much domain knowledge. It is clearly demonstrated that our method is much more effective. A possible reason is that the one in~\cite{padron2019improved} cannot handle well the impact on subsequent operations when scheduling for the early or current ones. While ours is able to simultaneously optimize any combination of variables with respect to different operations, vehicles and so on. On the other hand, the GA method needs to maintain a population of solutions, with higher demands in computational time and space (memory usage), especially for large-scale instances. Meanwhile, the design of crossover and mutation operators requires much domain knowledge to satisfy the complex constraints and ensure the feasibility of offspring. We circumvent such issues by directly using the off-the-shelf solver and learning to decompose the computation-heavy problem into a series of small sub-problems.


\begin{table}[!t]  
  \newcommand{\tabincell}[2]{\begin{tabular}{@{}#1@{}}#2\end{tabular}}
  \renewcommand{\arraystretch}{1.3}
  \renewcommand\tabcolsep{1.2pt}
  \caption{Results with OR-Tools.}
  \label{table_4}
  \centering
  \footnotesize
	\begin{tabular}{l rr rr rr}
		\toprule
	    & \multicolumn{2}{c}{AGH20} & \multicolumn{2}{c}{AGH50} & \multicolumn{2}{c}{AGH100} \\
		\cmidrule(lr){2-3}\cmidrule(lr){4-5}\cmidrule(lr){6-7}
			Methods & Obj. & Gap & Obj. & Gap & Obj. & Gap \\ 
			\midrule
			OR-Tools & \textbf{80073} & \textbf{4.07\%} & \textbf{150835} & \textbf{3.12\%} & 323762 & 30.47\% \\
			Random & 135183 & 75.86\% & 261293 & 78.77\% & 466692 & 88.29\% \\
			Vehicle-random & 88205 & 14.34\% & 159680 & 9.15\% & 313815 & 26.63\% \\
            Vehicle-worst & 88413 & 14.85\% & 184750 & 26.33\% & 502614 & 102.82\% \\
            Vehicle-worst-random & 88208 & 14.53\% & 162148 & 10.81\% & 352973 & 42.20\% \\
            IL-sample(Ours) & 86479 & 12.27\% & 151488 & 3.59\% & \textbf{300160} & \textbf{20.97\%} \\
		\bottomrule          
	\end{tabular}
	\vspace{-4mm}
\end{table}

\subsection{Learned LNS vs Heuristic Counterparts}\label{exp-3}
In our method, rather than manually designing complex operators, we attempt to automatically learn destroy operators with imitation learning, the demos of which are selected among multiple solutions by a heuristic, i.e., Vehicle-random LNS. We wonder that whether the learned destroy operator, based on the GCN representation of problem instances and solving states, could achieve better performance than the corresponding heuristic one. To verify this point, we compare the learning based LNS with Vehicle-random LNS, which is used as the provider of demos in imitation learning and also has shown better performance than other LNS heuristics on AGH20, AGH50 and AGH100 in Table \ref{table_3}. As discussed in section \ref{subsec:utilization}, we can deploy the learned destroy operator in two more ways besides the IL-sample used in section~\ref{exp-2}, i.e., IL-sampleA and IL-sampleD. For fair comparisons, we also apply Vehicle-random destroy operator in similar manners as follows, 1) Vehicle-random-sampleA, which applies Vehicle-random by using flexible destroy degree in each iteration, i.e., selecting a subset of variables with the percentage sampled from a range\footnote{We set it to $[0.2, 0.7]$, $[0.2, 0.5]$, $[0.2, 0.4]$ for AGH20, AGH50, AGH100 respectively, to render sub-MILPs tractable for each problem size and also prevent infeasible solutions.}; 2) Vehicle-random-sampleD, which is applied on top of Vehicle-random destroy operator, and imposes that the selected subsets between consecutive LNS iterations are disjoint. For IL-sampleA and IL-sampleD, we directly use the trained model of IL-sample with their respective sampling strategies in the deployment.
We evaluate the three different deployments for both the learned LNS and Vehicle-random LNS with 10 iterations on testing instances of all problem sizes. Fig.~\ref{fig_4} shows the average objective values by each method against iterations. 


\begin{table}[!t]
  \newcommand{\tabincell}[2]{\begin{tabular}{@{}#1@{}}#2\end{tabular}}
  \renewcommand{\arraystretch}{1.3}
  \renewcommand\tabcolsep{0.5pt}
  \caption{Results on AGH200 (450k/4000k).}
  \label{table_200}
  \centering
  \footnotesize
  \begin{tabular}{l|c|c|c|rrc}
  \toprule
   \multirow{2}*{Method} & \multirow{2}*{Mipgap} & \multirow{2}*{Timelimit} & Destroy & \multirow{2}*{Obj.} & \multirow{2}*{Gap} & \multirow{2}*{Time}\\
   &&& degree & &&\\
  \hline
   CPLEX & 1e-04 & 2h & -- & 1322757 & 152.72\% & 2h\\
   Decomposition & 1e-04 & 2.2h & -- & 663254 & 26.73\% & 2.2h\\
   Random & 0.2 & 30m & 0.2 & 1321166 & 152.41\% & 2h\\
   Vehicle-random & 0.2 & 30m & 0.2 & 666801 & 27.33\% & 2h\\
   Vehicle-worst & 0.2 & 30m & 0.2 & 1080257 & 106.66\% & 2h\\
   Vehicle-worst-random & 0.2 & 30m & 0.2 & 741957 & 41.27\% & 2h\\
   IL-sample (Ours) & 0.2 & 30m & 0.2 & \textbf{652405} & \textbf{24.59\%} & 2h\\
   \hline\hline
   OR-Tools & -- & 2h & -- & 1288353 & 146.28\% & 2h\\
   Random & -- & 30m & 0.2 & 1318479 & 151.97\% & 2h\\
   Vehicle-random & -- & 30m & 0.2 & 1084422 & 107.24\% & 2h\\
   Vehicle-worst & -- & 30m & 0.2 & 1120329 & 114.09\% & 2h\\
   Vehicle-worst-random & -- & 30m & 0.2 & 1087186 & 107.76\% & 2h\\
   IL-sample (Ours) & -- & 30m & 0.2 & \textbf{1015181} & \textbf{94.05\%} & 2h\\
  \bottomrule
  \end{tabular}
\end{table}

In general, we observe from Fig. \ref{fig_4} that IL-sampleA consistently achieves better solutions than the three different heuristic LNS on all problem sizes. It means that we can learn a more effective destroy operator from the simple Vehicle-random LNS, which guides LNS to search better solutions more efficiently. Also, the three deployments of the learned policy are better than or on par with their heuristic counterparts, which are summarized as below,

\noindent\textbf{IL-sample vs Vehicle-random.} With the same destroy degree (i.e., $\mathcal{D}_d=0.4$), the results indicate that the learned policy can select more suitable variables in each iteration, which generally leads to more efficient improvement of the solutions for problem instances of different sizes.

\noindent\textbf{IL-sampleA vs Vehicle-random-sampleA.} Both methods dynamically select a subset of variables to be optimized in each iteration. However, IL-sampleA significantly outperforms Vehicle-random-sampleA, which suggests that the learned policy, based on GCN representation of the instance and solving state, is able to construct desirable sub-MILPs that lead to better solutions along the solving process. 

\noindent\textbf{IL-sampleD vs Vehicle-random-sampleD.} When the selected variables between consecutive iterations are exclusive, IL-sampleD is on par with its counterpart for all problem sizes. It is fairly acceptable, given the fact that we do not intentionally involve any disjoint sample during training, since it directly uses the trained model of IL-sample.

In addition, we further evaluate the trends of solution improvement along the runtime for each deployment. 
Specifically, we run all these methods with the same time as in Table \ref{table_3}, i.e., 1m, 5m, 30m for AGH20, AGH50, AGH100, respectively. Fig. \ref{fig_5} shows the incumbent solutions against the runtime. We observe that the deployments of the learned destroy operator can generally retain the incumbents along the solving process for all instances, which means they are more efficient than heuristic destroy operators to achieve better solutions. It is also revealed that there is no single deployment dominating all the rest. On AGH20, IL-sample and IL-sampleD alternately achieve incumbents at the early stage, while finally IL-sampleA attains the lowest objective value. On AGH50, IL-sampleD retains the incumbent solution along almost the whole solving process. On AGH100, IL-sample, IL-sampleA and IL-sampleD yield the incumbent in early, middle and late stages, respectively. 


Considering that the policy network is only trained with demos of Vehicle-random on AGH20, the above results indicate that our method can deliver a more effective destroy operator than the heuristic counterpart, and the deployments with different sampling strategies may also further boost the performance, i.e., adaptive or disjoint sampling.

\begin{figure*}[!t]
\centering
\begin{tabular}{@{}c@{}c@{}}
  {\includegraphics[width=3in,height=1.8in]{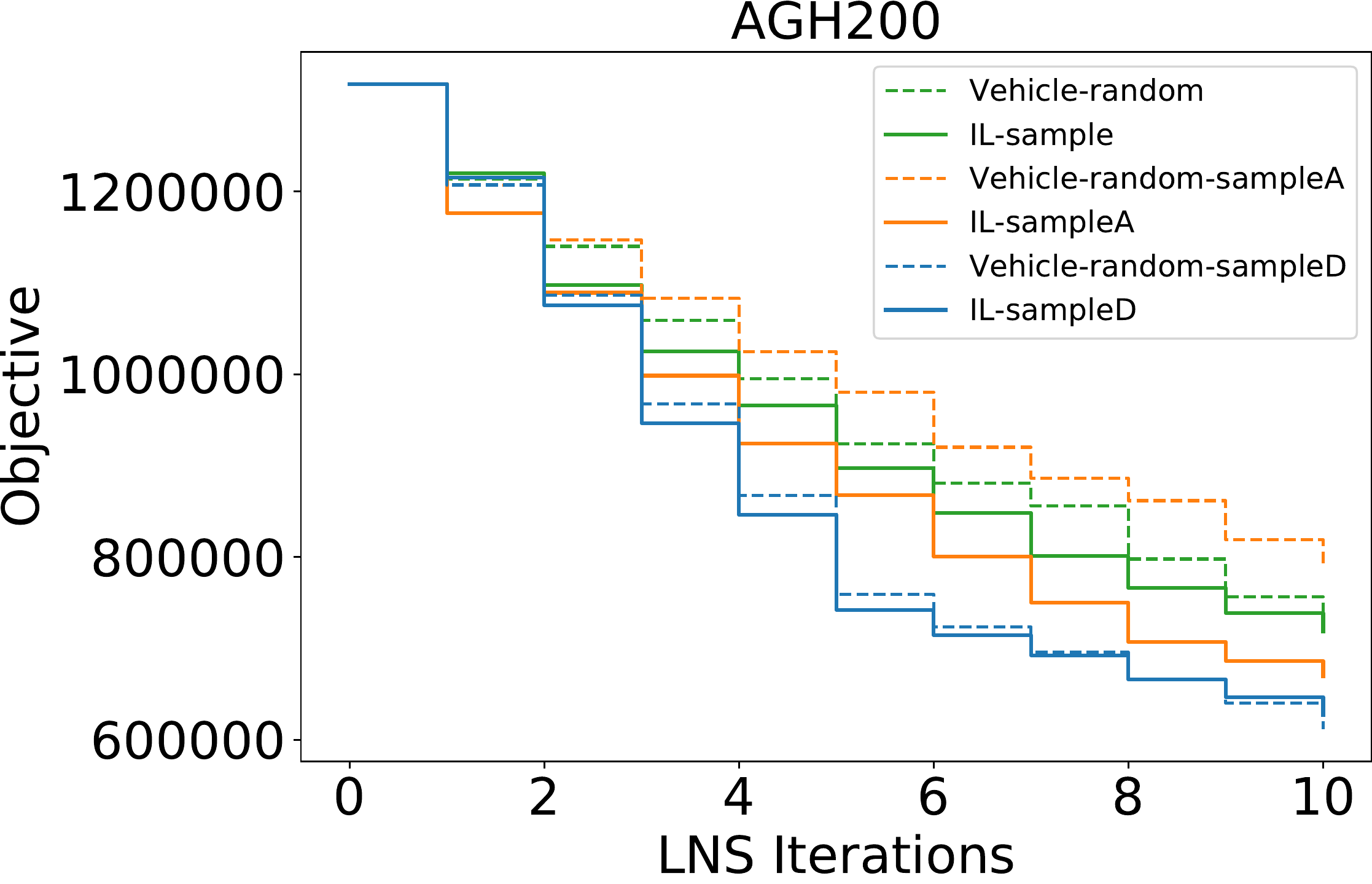}}&\hspace{10mm}
  {\includegraphics[width=3in,height=1.8in]{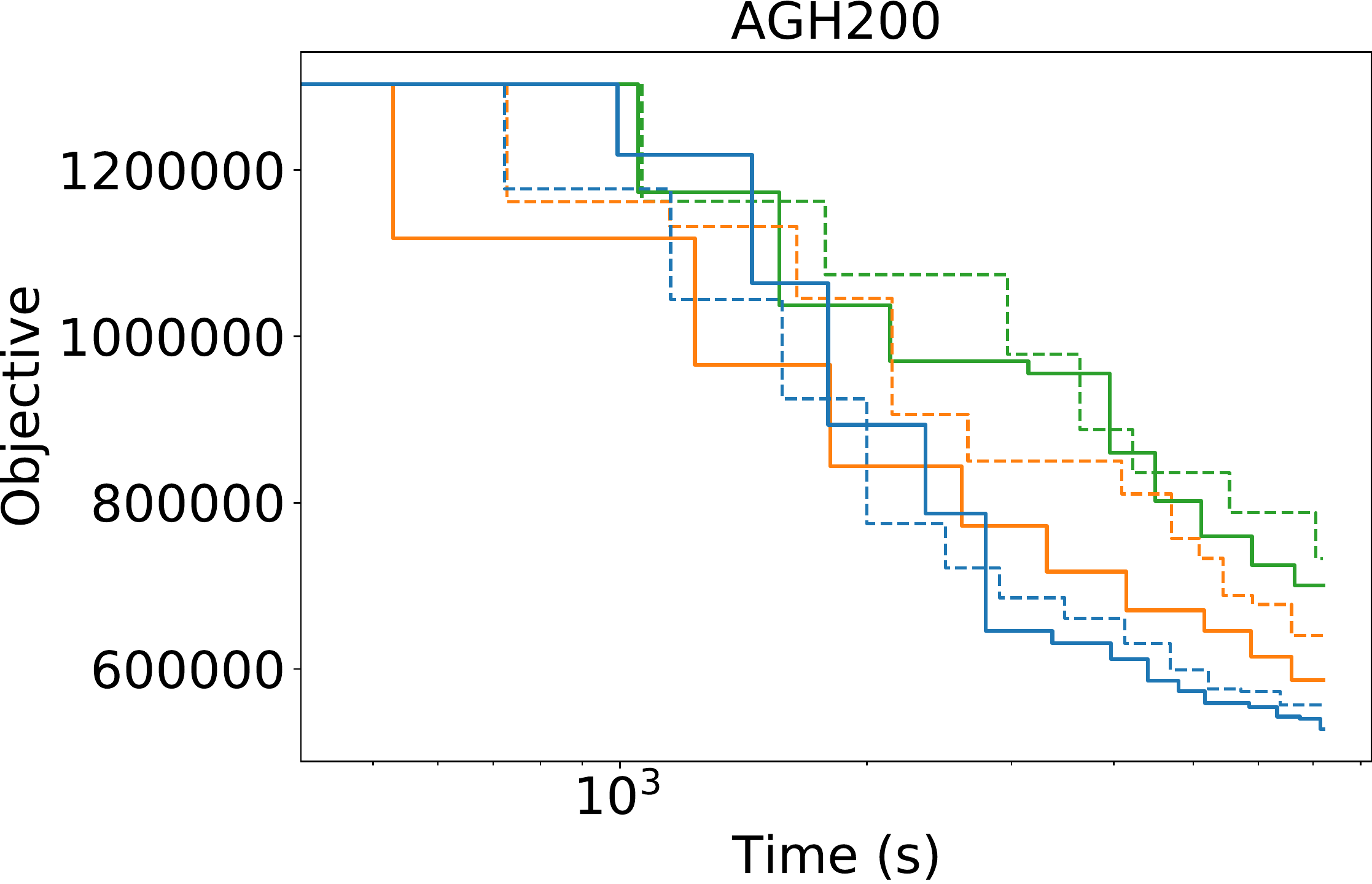}}
\end{tabular}
\caption{Comparison between the learned LNS and heuristic counterparts (left: the objective values of incumbents against iterations; right: the objective values of incumbents against runtime).}
\label{fig_6} 
\vspace{-2mm}
\end{figure*}

\subsection{Versatility Analysis}\label{exp-4}
One desirable potential of the proposed method is that we can integrate it with any off-the-shelf solver, which saves considerable efforts in manually designing the repair operator. To verify this merit of versatility, we replace CPLEX with OR-Tools (v8.1), i.e., a popular open source solver, as the repair operator in the LNS methods. In specific, we train the policy network on instances of AGH20, and test the learned LNS and LNS heurisitics on instances of AGH20, AGH50 and AGH100 as similarly done for CPLEX in section \ref{exp-2}. We keep the destroy degree of each method the same as the ones in Table \ref{table_3}, and ignore the mipgap since it is not explicitly accessible in OR-Tools. For all LNS methods, we set the time limit in each iteration to 25s, 120s and 500s on AGH20, AGH50 and AGH100, and the total runtime is set to 4m, 30m and 1.5h, respectively. We record the average objective values and the primal gap over the testing set, and display them in Table \ref{table_4}. Note that we only consider the deployment of IL-sample as the learning method for demonstration purpose.



We observe that IL-sample still works fairly well as a learning scheme to guide OR-Tools. Compared to other LNS baselines which are generally inferior to OR-Tools (except Vehicle-random on AGH100), we can see that IL-sample performs consistently better on all problem sizes. On AGH20, IL-sample significantly outperforms other LNS heuristics, and on AGH50 it becomes comparable to the standalone OR-Tools. On AGH100, IL-sample achieves considerably smaller objective value and gap than all other methods, revealing the beneficial property of our method on the large-scale problem. Therefore, the versatility of our framework is well verified, which can yield competitive performance even if we adopt different solvers as the repair operator.


\subsection{Scalability Analysis}
\label{exp-5}
Our method has already demonstrated desirable scalability property, however, we would like to further apply it to solve much larger instances, i.e., AGH200. Although the most recent literature on applying deep (reinforcement) learning to solve CVRP or CVRPTW~\cite{hottung2019neural,gao2020learn,xu2021reinforcement,xin2020multi} also deal with large instances, the problem settings in those works are quite standard or simple, which are far easier than the ones in this paper. Therefore, we are keen on evaluating the performance of our method on AGH200. To this end, we generate 20 instances with 200 flights, and solve them by solvers (i.e. CPLEX and OR-Tools), the decomposition method~\cite{padron2019improved} following previous setting, and LNS methods with  CPLEX and OR-Tools as the repair operators, respectively\footnote{We did not compare with GA here since it consumes too much memory to keep the population for AGH200, and fails in our device.}. For all these methods, we set the time limit and total runtime to 30m and 2h, respectively. In specific, we set mipgap to 0.2 for CPLEX in each iteration, and set the same destroy degree $\mathcal{D}_d=0.2$ for all LNS methods to keep sub-problems tractable. For the learned LNS, we still use the one trained on AGH20 and only consider fix-sized sampling strategy for inference, i.e., IL-sample. Then, we compute the average objective and primal gap over the tested instances for each method. All results are displayed in Table~\ref{table_200}, where the upper and lower half show the results for methods associated with CPLEX and OR-Tools, respectively. It is clear that no matter which solver is used as a repair operator, our method significantly outperforms other baselines, with the smallest average objective and gap. Most of the LNS methods demonstrated salient advantages over CPLEX and OR-Tools, since they can iteratively search promising solutions by solving sequential sub-problems. The decomposition method with slightly longer runtime surpasses LNS heuristics, but it is still far inferior to IL-sample (Ours). The superiority of our method comes from the fact that, in comparison with LNS heuristics, it can effectively guide LNS by determining more appropriate sub-MILPs that lead to better solutions. It is also suggested that the policy trained for small size of instances can generalize well to large ones, which may result from the desirable scaling property of GCN for learning the underlying representation of MILP.

To further verify that the learned policy is superior to the heuristic on AGH200, we compare their different deployments as did in section \ref{exp-3}. Specifically, we evaluate the three deployments on test instances with 10 iterations, and the curves of average objective for each deployment are shown in the left panel of Fig.~\ref{fig_6}. As shown, IL-sample is able to improve the incumbent more efficiently than Vehicle-random, which provides the demos for training. We also find that though we do not involve other schemes (i.e. adaptive and disjoint sampling) during training, the results achieved by IL-sampleA and IL-sampleD are still better than or on par with their counterparts, i.e., Vehicle-random-sampleA and Vehicle-random-sampleD, respectively. Furthermore, we also present curves to show the objective against 2h runtime. From the right panel of Fig.~\ref{fig_6}, we observe that the deployments of the learned LNS generally outperform their heuristic counterparts, which again indicates that the LNS with the learned destroy operator is more effective in boosting the solution quality.


\begin{figure}[!t]
  \centering
  \includegraphics[scale=0.4]{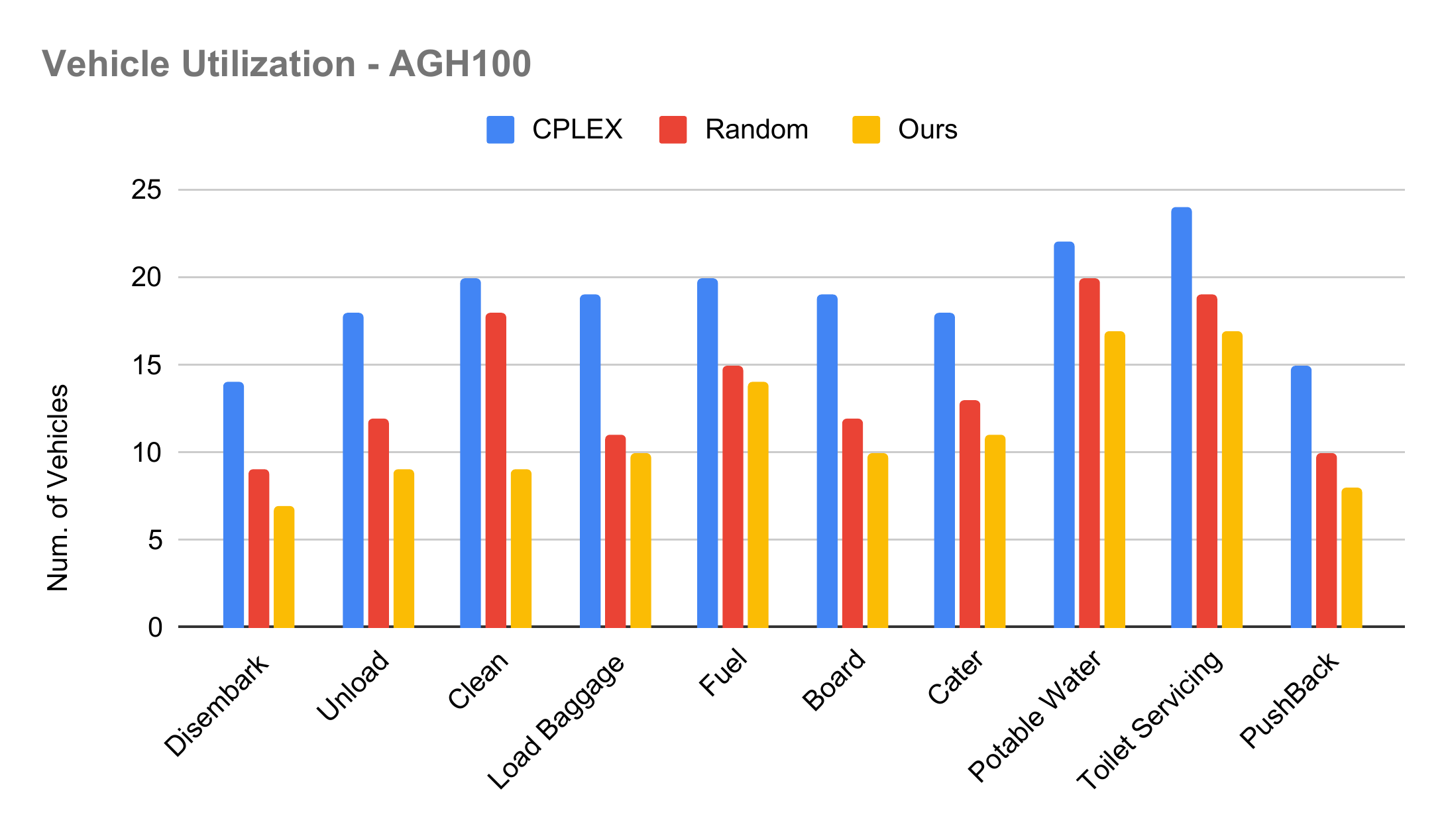}
  \caption{The average number of occupied vehicles in different fleets.}
  \label{fig_visu}
  \vspace{-2mm}
\end{figure}

\subsection{Effect on Vehicle Utilization}

Despite that we merely set the objective to explicitly minimize the total tour length of vehicles, we empirically observe that our method is able to concurrently achieve good performance in terms of the vehicle utilization. To manifest this point, we solve AGH100 instances in the testing set by CPLEX, Random LNS and our method as did in section \ref{exp-2}, and record the average number of occupied vehicles in every fleet over all the instances. The results are displayed in~Fig.~\ref{fig_visu}. It is clear that our method attains a better solution which employs significantly less vehicles in each fleet, in comparison with CPLEX and Random LNS. Our method implicitly enables a reasonable vehicle utilization in each fleet, serving all aircraft on time with a small number of vehicles. Considering that an airport generally possesses a limited number of vehicles for AGH, our method is more suited to make the best usage of this critical resource.

We would like to note that this paper mainly contributes to a learning based framework to solve AGH with multiple heterogeneous operations, which has been verified superior to various baselines. Nevertheless, the adaptation of our method to AGH with different objective functions is straightforward, by modelling them with an off-the-shelf solver and substituting the objective in this paper. In the future, we will apply our method to solve AGH with other potential objectives to further verify its effectiveness, such as explicitly minimizing vehicle utilization~\cite{zhao2019bi} and maximizing safety to prevent accidents~\cite{shvetsov2021analysis,shvetsova2021ensuring}.


\section{Conclusions and Future Works}
\label{sec:conclusion}
To our knowledge, we are the first one to leverage learning based LNS framework to solve the complex vehicle routing problem in AGH with multiple operations and hundreds of flights. In this framework, we integrate imitation learning and GCN to guide the destroy operator to automatically select variables in each iteration, and adopt the off-the-shelf solver as the repair operator to reoptimize the selected variables, which circumvents substantial human efforts for designing the rules in conventional LNS heuristics. Results demonstrate that our method has a strong capability in delivering high-quality solutions compared to the exact solvers and conventional heuristics, with desirable properties in versatility and scalability. Hence, the proposed method has a favorable potential that we could fast deploy an off-the-shelf solver and automatically raise its performance without much manual work or domain knowledge.

As mentioned, this work is an early attempt on applying learning based method to tackle the complex vehicle routing problem in airports. In future, we will, 1) investigate deep reinforcement learning to learn the destroy operators, and test our method on other datasets; 2) study how to enable the destroy operator to automatically decide the number of variables (i.e., the size of subproblems) to be optimized;
3) take into account the safety factor to avoid potential accidents in the organization of traffic; 
4) extend the policy network to optimize multiple objectives in the MILP model.

\section*{Acknowledgements}
This research was conducted in collaboration with Singapore Telecommunications Limited and supported by the Singapore Government through the Industry Alignment Fund - Industry Collaboration Projects Grant. It was also supported by the National Natural Science Foundation of China (Grant No. 62102228), and the Natural Science Foundation of Shandong Province (Grant No. ZR2021QF063).

\bibliographystyle{IEEEtran}
\bibliography{IEEEabrv, camera_ready}


%





\ifCLASSOPTIONcaptionsoff
  \newpage
\fi

\begin{IEEEbiography}[{\includegraphics[width=1in,height=1.25in,clip,keepaspectratio]{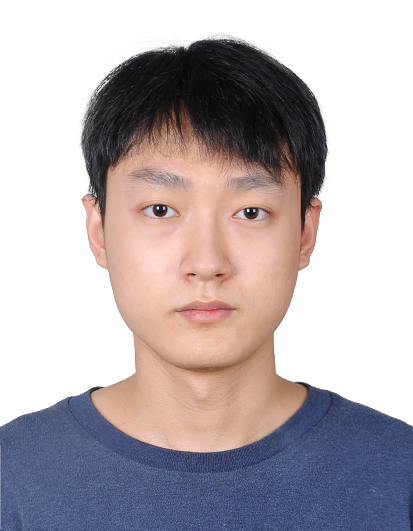}}]{Jianan Zhou}
received the B.Eng. degree in software engineering from Northeastern University, Shenyang, China, in 2019, and the M.Sc. degree in artificial intelligence from Nanyang Technological University, Singapore, in 2021. He is currently pursuing the Ph.D. degree with the School of Computer Science and Engineering, Nanyang Technological University, Singapore. His research interest includes machine learning with combinatorial optimization problems.
\end{IEEEbiography}

\begin{IEEEbiography}[{\includegraphics[width=1in,height=1.25in,clip,keepaspectratio]{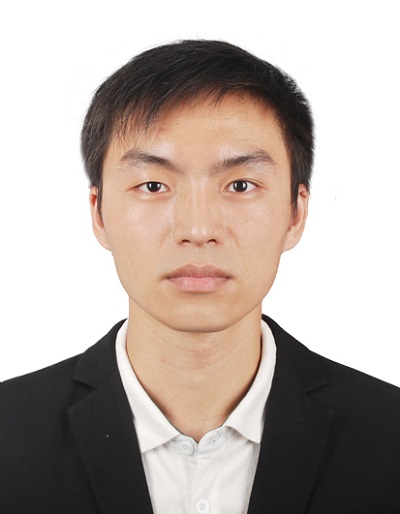}}]{Yaoxin Wu}
received the B.Eng degree in traffic engineering from Wuyi University, Jiangmen, China, in 2015, the M.Eng degree in control engineering from Guangdong University of Technology, Guangzhou, China, in 2018, and the Ph.D. degree in computer science from
Nanyang Technological University, Singapore, in 2023. He was a Research Associate with the Singtel Cognitive and Artificial Intelligence Lab for Enterprises (SCALE@NTU). He joins the
Department of Information Systems, Faculty of
Industrial Engineering and Innovation Sciences, Eindhoven University of Technology, as an Assistant Professor. His research interests include combinatorial optimization, integer programming and deep learning.
\end{IEEEbiography}

\begin{IEEEbiography}[{\includegraphics[width=1in,height=1.25in,clip,keepaspectratio]{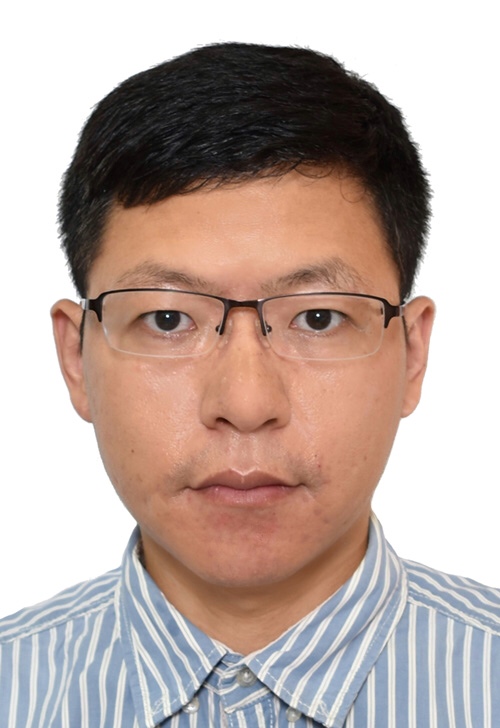}}]{Zhiguang Cao}
received the Ph.D. degree from Interdisciplinary Graduate School, Nanyang Technological University. He received the B.Eng. degree in Automation from Guangdong University of Technology, Guangzhou, China, and the M.Sc. in Signal Processing from Nanyang Technological University, Singapore, respectively. He was a Research Fellow with the Energy Research Institute @ NTU (ERI@N), a Research Assistant Professor with the Department of Industrial Systems Engineering and Management, National University of Singapore, and a Scientist with the Agency for Science Technology and Research (A*STAR), Singapore. He joins the School of Computing and Information Systems, Singapore Management University, as an Assistant Professor. His research interests focus on learning to optimize (L2Opt).
\end{IEEEbiography}

\begin{IEEEbiography}[{\includegraphics[width=1in,height=1.25in,clip,keepaspectratio]{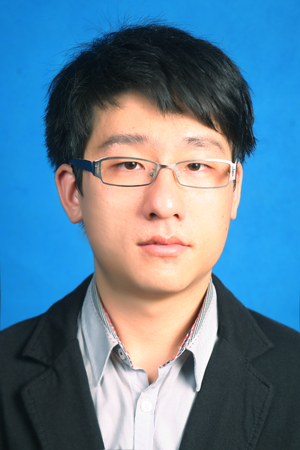}}]{Wen Song}
received the B.S. degree in automation and the M.S. degree in control science and engineering from Shandong University, Jinan, China, in 2011 and 2014, respectively, and the Ph.D. degree in computer science from Nanyang Technological University, Singapore, in 2018. He was a Research Fellow with the Singtel Cognitive and Artificial Intelligence Lab for Enterprises (SCALE@NTU). He is currently an Associate Research Fellow with the Institute of Marine Science and Technology, Shandong University. His current research interests include artificial intelligence, planning and scheduling, multi-agent systems, and operations research.
\end{IEEEbiography}

\begin{IEEEbiography}[{\includegraphics[width=1in,height=1.25in,clip,keepaspectratio]{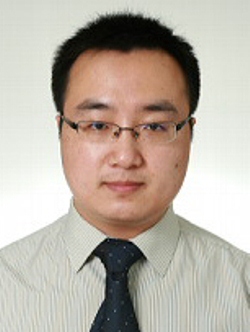}}]{Jie Zhang}
received the Ph.D. degree from the Cheriton School of Computer Science, University of Waterloo, Canada, in 2009. He is currently a Professor with the School of Computer Science and Engineering, Nanyang Technological University, Singapore. He is also a Professor at the Singapore Institute of Manufacturing Technology. During his Ph.D. study, he held the prestigious NSERC Alexander Graham Bell Canada Graduate Scholarship rewarded for top Ph.D. students across Canada. He was also a recipient of the Alumni Gold Medal at the 2009 Convocation Ceremony. The Gold Medal is awarded once a year to honour the top Ph.D. graduate from the University of Waterloo. His papers have been published by top journals and conferences and received several best paper awards.
\end{IEEEbiography}

\begin{IEEEbiography}[{\includegraphics[width=1in,height=1.25in,clip,keepaspectratio]{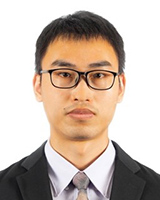}}]{Zhenghua Chen (Member, IEEE)}
received the
B.Eng. degree in mechatronics engineering
from the University of Electronic Science and
Technology of China, Chengdu, China, in 2011,
and the Ph.D. degree in electrical and electronic
engineering from Nanyang Technological University (NTU), Singapore, in 2017.
He is working with NTU as a Research Fellow. He is currently a Scientist with Institute for
Infocomm Research (I2R), Agency for Science, Technology and Research (A*STAR), Singapore. His
research interests include sensory data analytics, machine learning,
deep learning, and transfer learning and related applications.
\end{IEEEbiography}








\end{document}